\documentclass[conference]{IEEEtran}
\usepackage{newtxtext,newtxmath}   

\ifCLASSINFOpdf
\else
\fi
\usepackage{graphicx}

\usepackage[pass]{geometry}

\usepackage{xcolor}
\usepackage{caption}
\usepackage{subcaption}
\usepackage{tikz}
\usetikzlibrary{shapes.geometric, arrows}
\usepackage{graphicx}  
\usepackage{caption}   
\usepackage{adjustbox} 

\usepackage{algorithm}
\usepackage{algorithmicx}
\usepackage{algpseudocode}
\usepackage{amsmath}
\usepackage{amsfonts}
\usepackage{graphicx}
\usepackage{float}
\usepackage{multirow}
\usepackage{array}
\usepackage{enumitem}
\usepackage{booktabs}
\usepackage{multirow}
\usepackage{subcaption} 
\usepackage{comment}
\usepackage{hyperref}
\usepackage{tabularx}   
\usepackage{placeins}
\begin{document}

%

\title{GuardNet: Graph-Attention Filtering for Jailbreak Defense in Large Language Models}



\author{\IEEEauthorblockN{Javad Forough}
	\IEEEauthorblockA{Imperial College London\\
		j.forough@imperial.ac.uk}
	\and
	\IEEEauthorblockN{Mohammad Maheri}
	\IEEEauthorblockA{Imperial College London\\
		m.maheri23@imperial.ac.uk}
	\and
	\IEEEauthorblockN{Hamed Haddadi}
	\IEEEauthorblockA{Imperial College London\\
		h.haddadi@imperial.ac.uk}}


%


\maketitle

\begin{abstract}
Large Language Models (LLMs) are increasingly susceptible to jailbreak attacks, which are adversarial prompts that bypass alignment constraints and induce unauthorized or harmful behaviors. These vulnerabilities undermine the safety, reliability, and trustworthiness of LLM outputs, posing critical risks in domains such as healthcare, finance, and legal compliance. In this paper, we propose \textbf{GuardNet}, a hierarchical filtering framework that detects and filters jailbreak prompts prior to inference. GuardNet constructs structured graphs that combine sequential links, syntactic dependencies, and attention-derived token relations to capture both linguistic structure and contextual patterns indicative of jailbreak behavior. It then applies graph neural networks at two levels: (i) a \emph{prompt-level filter} that detects global adversarial prompts, and (ii) a \emph{token-level filter} that pinpoints fine-grained adversarial spans. Extensive experiments across three datasets and multiple attack settings show that GuardNet substantially outperforms prior defenses. It raises prompt-level F$_1$ scores from 66.4\% to 99.8\% on \textsc{LLM-Fuzzer}, and from 67–79\% to over 94\% on \textsc{PLeak} datasets. At the token level, GuardNet improves F$_1$ from 48–75\% to 74–91\%, with IoU gains up to +28\%. Despite its structural complexity, GuardNet maintains acceptable latency and generalizes well in cross-domain evaluations, making it a practical and robust defense against jailbreak threats in real-world LLM deployments.
\end{abstract}


%
\IEEEpeerreviewmaketitle

\section{Introduction}  
Large Language Models (LLMs) have become central to a wide range of applications, powering systems in domains such as education \cite{voultsiou2025systematic}, healthcare \cite{shool2025systematic}, finance \cite{li2023large}, law \cite{siino2025exploring}, and customer support \cite{soni2023large}. Their ability to understand and generate human-like text has enabled automation of complex tasks such as legal reasoning, clinical triage, financial analysis, and policy drafting. However, this general-purpose capability also makes LLMs vulnerable to misuse. In particular, LLMs are highly susceptible to prompt-based adversarial attacks, especially \textit{jailbreak prompts} \cite{yi2024jailbreak, wei2023jailbroken}, which are carefully engineered inputs designed to bypass alignment constraints and elicit unauthorized or harmful responses. These prompts can override safety protocols, suppress built-in refusals, or trick models into producing content that violates ethical, legal, or security boundaries.

The danger posed by jailbreak attacks extends beyond content safety. They can be used to extract hidden model behaviors, circumvent content filters, or even exfiltrate proprietary information embedded in model responses. Such attacks not only compromise the intended behavior of LLMs but also erode public trust in their reliability and safety, especially in applications where strict adherence to operational, legal, or ethical guidelines is non-negotiable.

Recent work has highlighted the sophistication of jailbreak attacks~\cite{pleak2024, yu2024llm, deng2024masterkey, dong2025fuzz, mehrotra2024tree}, demonstrating the ability to exploit both syntactic patterns and semantic cues in prompts. These methods leverage structural perturbations, adversarial phrase injection, and iterative query refinements to manipulate model outputs. Crucially, such attacks operate entirely at the input level and are conducted in a black-box manner, requiring no access to model weights or internals, making them highly stealthy, transferable, and difficult to detect using traditional defenses.

As LLMs continue to be deployed in increasingly sensitive and mission-critical settings, it is imperative to develop robust, adaptive defenses that can reliably detect and mitigate jailbreak attempts at inference time. Supervised machine learning has been successfully applied for attack detection in diverse domains, including edge computing \cite{forough2023anomaly, wei2025toward}, network intrusion detection \cite{jha2025netprobe, wei2023xnids}, and financial fraud prevention \cite{forough2022sequential, faisal2024fraud}. These successes highlight its capacity to learn discriminative patterns from labeled threat data, motivating its adaptation to the emerging challenge of jailbreak detection in LLMs.

In this paper, we propose GuardNet, a hierarchical filtering framework based on supervised machine learning, designed to detect and suppress jailbreak prompts at inference time. GuardNet constructs hybrid token graphs that integrate syntactic dependency structures with multi-head attention patterns, capturing both linguistic and contextual patterns indicative of jailbreak behavior. It then applies Graph Neural Networks (GNNs) at two levels:
\begin{enumerate}
    \item \textit{Prompt-Level Filtering}: A coarse-grained GNN that models global contextual coherence across the full prompt, identifying high-level adversarial intent.
    \item \textit{Token-Level Filtering}: A fine-grained GNN that detects and localizes malicious spans via attention-based anomaly scoring and embedding shift analysis.
\end{enumerate}

Unlike existing rule-based or shallow classifiers, GuardNet provides a modular, real-time defense that does not require retraining or modifying the target LLM. It operates as a preprocessing layer, enabling integration with existing pipelines with minimal overhead.

Our contributions are as follows:
\begin{itemize}
    \item We introduce a hierarchical graph-based adversarial filtering framework that fuses sequential links, attention-derived token relations, and syntactic dependencies to form rich, multi-view representations of input prompts.

    \item We develop a prompt-level graph neural network that detects anomalous global intent by modeling contextual coherence and structural inconsistencies indicative of jailbreak behavior.

    \item We propose a token-level graph neural network that localizes adversarial spans using attention-based anomaly scoring and contextual embedding shifts, enabling fine-grained threat detection.

    \item We perform extensive evaluations on two recent jailbreak benchmarks (PLeak and LLM-Fuzzer) across diverse domains, demonstrating that GuardNet not only outperforms existing defenses in detection accuracy and robustness, but also operates efficiently as a pre-inference module suitable for real-time deployment without modifying the underlying LLM.
\end{itemize}

The remainder of this paper is organized as follows. In Section~\ref{sec:related_work}, we review related work on jailbreak attacks and existing defenses, highlighting the limitations that motivate our design. In Section~\ref{sec:problem_statement}, we formally define the problem and our objectives, and in Section~\ref{sec:threat_model}, we describe the threat model and assumptions.  In Section~\ref{sec:framework}, we present the GuardNet framework, detailing its graph-based architecture and hierarchical filtering components. Section~\ref{sec:evaluation} describes our experimental setup and evaluation methodology, followed by the quantitative and qualitative results. Finally, Section~\ref{sec:discussion} discusses the advantages and limitations of the proposed framework, and Section~\ref{sec:conclusion} concludes the paper and outlines directions for future research.

\section{Related Works}
\label{sec:related_work}
\subsection{Security Challenges in Large Language Models}

Large Language Models (LLMs) have demonstrated remarkable capabilities across a wide range of language tasks, yet their open-ended, prompt-driven nature leaves them vulnerable to a growing class of \textit{jailbreak} attacks. These attacks manipulate the input prompt to elicit unintended or unsafe behavior from the model, often bypassing alignment constraints without altering model parameters. Critically, such attacks are typically black-box, requiring only query access to the model and no knowledge of its architecture or weights. The most concerning variants are those that either extract hidden system information (e.g., prompt leakage), or coerce the model into violating its safety policies via contextual manipulation. As the sophistication of these threats grows, there is an urgent need for robust input-side defenses that can detect and mitigate adversarial intent prior to inference. In the following, we will discuss some of the key jailbreak attacks and different related existing defenses that motivate the design of GuardNet.

\subsubsection{\textbf{AutoDAN: Automated Jailbreaks via Discrete Optimization}}
AutoDAN~\cite{liu2023autodan} introduces a black-box adversarial attack that automates the discovery of jailbreak prompts using discrete optimization techniques. By framing jailbreak crafting as a sequence-level search problem, AutoDAN leverages a compact surrogate model to iteratively mutate prompts that evade safety filters while preserving coherence. Its efficiency stems from a learnable discrete action space that significantly reduces query cost. While similar in spirit to LLM-Fuzzer, AutoDAN's optimization-driven strategy is highly effective at generating transferable and stealthy adversarial prompts. Its success demonstrates the urgent need for proactive, input-side defenses capable of anticipating unseen perturbations.

\subsubsection{\textbf{PLeak: Prompt Leakage Attack}}

PLeak~\cite{pleak2024} is a black-box jailbreak attack designed to extract confidential system prompts from LLMs. It operates by evolving adversarial user prompts using gradient-free optimization (e.g., CMA-ES) to induce model completions that reveal parts of the hidden system prompt. This disclosure represents a serious threat, as system prompts often encode proprietary behaviors, ethical constraints, or usage policies. To evade naive output filtering mechanisms, PLeak further introduces randomized transformations of the extracted text (e.g., paraphrasing or synonym substitution), increasing robustness against sanitization.

\subsubsection{\textbf{LLM-Fuzzer}}

LLM-Fuzzer~\cite{yu2024llm} (also referred to as GPTFuzzer) systematically generates jailbreak prompts by learning patterns from a small number of seed jailbreaks. It constructs a fuzzing grammar and leverages mutation and recombination strategies to explore the space of adversarial inputs. These synthetic prompts are iteratively tested against the model to find candidates that bypass safety filters or alignment restrictions. Crucially, the method is model-agnostic and black-box, making it broadly applicable to different LLMs without requiring access to internal weights or training data. LLM-Fuzzer exposes the structural weaknesses in prompt-based alignment, demonstrating that relatively simple prompt mutations can reliably subvert system-level safeguards.

These attacks highlight an urgent need for proactive defenses capable of detecting adversarial intent directly from the input prompt, rather than relying on post hoc output filtering or restrictive fine-tuning. In this context, our proposed GuardNet framework addresses this gap by performing hierarchical analysis of both prompt structure and token-level semantics to detect and block such manipulations before inference.

\subsection{Existing Defenses and Their Limitations}

Traditional defense strategies against adversarial inputs suffer from notable limitations. Perplexity-based filtering fails to detect sophisticated prompts that maintain plausible surface-level fluency. Rule-based approaches, while fast, rely on static heuristics that can be easily circumvented by minor rewording or paraphrasing. To address the limitations of classical methods, recent research has explored data-driven techniques that capture distributional and structural irregularities in input text. Some of the most recent and notable approaches are described in the following sections.

\subsubsection{\textbf{JailGuard: Mutation-Based Divergence Detection}}
JailGuard~\cite{zhang2025jailguard} defends against jailbreak prompts by creating multiple slightly altered versions of each input using diverse text mutators such as synonym substitution, punctuation tweaks and paraphrasing, then analyzing how much the model’s outputs differ across these variants. The core intuition is that benign prompts yield consistent responses even under small perturbations, whereas adversarial inputs crafted to bypass alignment will produce unusually high variability. Because it treats the target model as a black box, JailGuard can be deployed without retraining or modifying model weights and it is compatible with any API-based service. However, this robustness requires multiple inference calls per prompt which increases both latency and API usage. Its effectiveness also depends on the predefined set of mutators and divergence thresholds, parameters that must be calibrated on known attack patterns and that may not generalize to novel or domain-specific manipulations 

\subsubsection{\textbf{ToxicDetector: Grey-Box Detection via Concept-Prompt Embeddings}}  
Liu et al.~\cite{liu2024efficient} present \emph{ToxicDetector}, a lightweight grey-box framework that leverages internal embedding features to detect toxic prompts in LLMs. Rather than treating the prompt text alone, ToxicDetector first generates a small set of “toxic concept” descriptions (e.g., “instructions for committing a crime”) via an LLM, then augments them to cover diverse semantic variants. During inference, for each user prompt and each concept, it extracts the embedding of the last token at every transformer layer and computes an element-wise product between the prompt’s and concept’s embeddings. These layer-wise products are concatenated into a single feature vector, which is classified by a shallow five-layer MLP. This design, however, requires grey-box access to internal embeddings, which makes it inapplicable to fully black-box APIs, and it depends on manually generated concept prompts and similarity thresholds, which can limit its adaptability to novel or evolving toxic scenarios.

\subsubsection{\textbf{TextDefense: Word Importance Dispersion for Global Detection.}}  
TextDefense~\cite{shen2025textdefense} is a model-agnostic framework that detects adversarial inputs by analyzing the dispersion of word importance scores. It builds on the insight that adversarial examples often perturb the semantic relevance of input tokens, leading to unstable or skewed importance distributions. By quantifying this dispersion through Word Importance Score Dispersion (WISD), TextDefense identifies inputs with abnormal influence patterns without modifying the underlying model. Although not specifically tailored for prompts, this method is effective at flagging anomalous input behavior at a coarse level. However, it cannot localize specific adversarial spans, limiting its precision in filtering complex, stealthy jailbreak prompts.

\subsubsection{\textbf{Dynamic Attention: Fine-Grained Detection via Attention Deviations.}}  
Dynamic Attention~\cite{shen2024improving} focuses on token-level detection by analyzing statistical anomalies in the attention maps of transformer models. It measures the deviation of each token’s attention distribution from expected norms derived from benign data, assigning anomaly scores that highlight tokens likely to be adversarial. This enables the detection of subtle manipulations such as trigger phrases or context reweighting. Although effective for localized threats, Dynamic Attention assumes static attention patterns and may struggle in cases with broader semantic manipulation. Moreover, it lacks explicit modeling of syntactic structure or higher-level context, making it susceptible to attacks that exploit grammatical consistency.

While each of these defenses contributes important techniques, they each address only one aspect of the problem. Mutation and divergence checks add extra queries and may fail on attacks outside their predefined transformations. Embedding-based methods depend on internal model access and manual concept engineering, limiting their applicability. Word-importance dispersion flags global irregularities but cannot pinpoint malicious spans. Attention-deviation techniques detect local anomalies without modeling higher-order structure. No single solution combines global intent detection, fine-grained localization, and structural understanding. This gap motivates a unified framework that integrates both linguistic structure and semantic context to defend against sophisticated prompt-based jailbreak attacks.

\subsection{Symbols and Notations}
We provide a summary of the key symbols and notations used throughout this paper in Table~\ref{tab:symbols}. These notations encompass the formulation of LLM-based prompting, adversarial query detection, GuardNet’s hybrid graph-based architecture, and relevant performance metrics. The table serves as a reference for the mathematical symbols introduced in Sections~\ref{sec:problem_statement} through~\ref{sec:evaluation}, covering both the structural components of the proposed model and the evaluation criteria.

\begin{table}[t]
  \footnotesize
  \caption{Symbols and notation used throughout the paper}
  \label{tab:symbols}
  \begin{tabularx}{\columnwidth}{@{}lX@{}}
    \toprule
    \textbf{Symbol} & \textbf{Description} \\ \midrule
    $V$                    & Sub-word vocabulary of the LLM \\
    $\mathcal{L}_{\max}$   & Maximum context length supported by the LLM \\
    $Y$                    & Open-ended textual output space \\
    $f_{\theta}\colon V^{\le \mathcal{L}_{\max}} \!\to\! Y$
                           & Instruction-tuned LLM with parameters~$\theta$ \\
    $s$                    & Confidential \emph{system prompt} (policy / exemplars) \\
    $u$                    & User-supplied query prompt \\
    $p = s \,\|\, u$       & Full composite prompt given to the LLM \\
    $r = f_{\theta}(p)$    & Model response to $p$ \\
    $\mathcal{A}$          & Set of jailbreak / adversarial queries \\
    $\mathcal{G} = (\mathcal{G}_{\mathrm{prompt}},\,\mathcal{G}_{\mathrm{token}})$
                           & GuardNet filter (prompt- and token-level) \\
    $m$                    & Binary mask of adversarial tokens returned by $\mathcal{G}_{\mathrm{token}}$ \\
    $\Delta$               & Maximum latency budget allowed for $\mathcal{G}$ \\
    $\tau_P$               & Decision threshold for prompt-level score \\
    $\tau_T$               & Decision threshold for token-level score \\
    $H$                    & Last-layer hidden states from Longformer encoder \\
    $d$                    & Hidden-state dimensionality (Longformer) \\
    $A$                    & Averaged self-attention matrix $(L{\times}L)$ \\
    $k$                    & Number of top-attention neighbours retained per token \\
    $w$                    & Local attention window size (Longformer) \\
    $L$                    & Token sequence length of the input prompt \\
    $E,\,V$                & Edge set and node set of the hybrid token graph \\
    $\alpha, \gamma$       & Focal-loss class weight and focusing factor \\
    $\mathrm{FPR},\,\mathrm{FNR}$
                           & False-positive / false-negative rates \\ \bottomrule
  \end{tabularx}
\end{table}

\section{Problem Statement}
\label{sec:problem_statement}

Consider an instruction-tuned large–language model (LLM) 
\[
f_{\theta}:\;\mathcal{V}^{\le L_{\max}}\longrightarrow\mathcal{Y},
\]
where $\mathcal{V}$ is the vocabulary, 
$L_{\max}$ the maximum context length and 
$\mathcal{Y}$ the open-ended textual output space.  
At inference time the application constructs a \emph{composite prompt}
\[
p \;=\; s \;\Vert\; u,
\qquad 
p\in\mathcal{V}^{\le L_{\max}},
\]
by concatenating a confidential system prompt $s$
(e.g.\ policy) and a user-supplied
query $u$.  
The model’s response is $r = f_{\theta}(p)$.

\vspace{4pt}
\noindent\textbf{Jailbreak definition.}\;  
We say that a query $u$ is a \emph{jailbreak} if, for some security
policy~$\mathcal{P}$, the combined prompt $p=s\Vert u$ causes the model to
violate~$\mathcal{P}$ either by
\emph{(i)} revealing any non-empty substring of~$s$
(\emph{confidentiality breach}) or
\emph{(ii)} producing disallowed content (\emph{integrity breach}).
Formally,
\[
u \in \mathcal{A} 
\;\;\Longleftrightarrow\;\;
\bigl[\,\exists\,\text{policy clause } C\in\mathcal{P}\!:
f_{\theta}(s\Vert u)\not\models C \bigr].
\]

Where $\mathcal{A}$ denotes the set of all jailbreak queries, $\mathcal{P}$ the deployment’s security policy (a collection of clauses), and $C\in\mathcal{P}$ an individual policy clause. The expression $f_{\theta}(s\Vert u)\not\models C$ means that the model’s response to the composite prompt fails to satisfy clause $C$.

\vspace{4pt}
\noindent\textbf{Objective.}\;  
Given only the user query $u$ (the system prompt~$s$ never leaves the
secure boundary), we seek a real-time filter
\[
G = (G_{\text{prompt}},\,G_{\text{token}})
      :\mathcal{V}^{\le L_{\max}}\longrightarrow
      \{0,1\}\;\times\;\{0,1\}^{\le|u|}
\]
such that

\begin{enumerate}[label=(\roman*),topsep=2pt,itemsep=1pt,leftmargin=14pt]
\item \textbf{Prompt-level decision.}  
      $G_{\text{prompt}}(u)=1$ iff $u\in\mathcal{A}$  
      (binary jailbreak detection).

\item \textbf{Token-level localisation.}  
      $G_{\text{token}}(u)=m$ returns a binary mask 
      $m\in\{0,1\}^{|u|}$ where $m_i=1$ marks the $i$-th token in
      $u$ as adversarial, enabling selective sanitisation:
      \[
      \tilde{u} 
      \;=\;
      \bigl\{\,u_i\mid m_i=0\bigr\},
      \qquad
      s\,\Vert\,\tilde{u} \;\;\text{is forwarded to } f_{\theta}.
      \]
\end{enumerate}

GuardNet fulfils these requirements by (i) constructing a hybrid graph that integrates sequential token-adjacency links, top‑k self-attention edges, and syntactic dependency arcs extracted from~$u$, and (ii) applying a graph neural network over this multi-view structure to realise the two detection objectives outlined above.

\section{Threat Model}
\label{sec:threat_model}

\paragraph{Adversary capabilities}
The attackers has black-box access to the deployed LLM
endpoint: they may submit an unlimited number of queries
$u^{(1)},u^{(2)},\dots$ and observe the textual responses
$r^{(1)},r^{(2)},\dots$ but gains no insight into internal parameters,
activations,~or GuardNet’s intermediate features.
They cannot modify the system prompt~$s$, the model weights
$\theta$, or the GuardNet code.

\paragraph{Attack goals}
The adversary seeks to craft a query $u$ that satisfies at least one
of the following:

\begin{enumerate}[label=(G\arabic*),itemsep=1pt,leftmargin=4\parindent]
\item \textbf{Prompt leakage.}\;  Elicit any substring of the hidden
      system prompt~$s$ (e.g.\ via PLeak).
\item \textbf{Policy violation.}\;  Coerce the model into producing
      content disallowed by the deployment policy
      (e.g.\ hate speech, instructions for illicit activities, \ldots).
\end{enumerate}

\paragraph{Attack surface}
The attacker may employ automated jailbreak frameworks
(e.g.\ LLM-Fuzzer, PLeak), gradient-free optimisation, paraphrasing,
or staged dialogue without constraints on length, vocabulary, or
format beyond the model’s maximum input size~$L_{\max}$.

\paragraph{Defender assumptions}
GuardNet is deployed \emph{in front of} the LLM.  
It sees each user query \emph{in plaintext} but never the private
system prompt~$s$ nor the model’s logits.
GuardNet can perform arbitrary computations on~$u$,
including third-party NLP parsing.

\paragraph{Security goal}  
Under the threat model described above, we evaluate both prompt‐level and token‐level detection via the F\textsubscript{1} score.  

At the prompt level:
\[
\begin{aligned}
  \mathrm{Precision}_{P}
    &= \frac{\bigl|\{\,u : G_{\mathrm{prompt}}(u)=1 \land u\in\mathcal{A}\}\bigr|}
           {\bigl|\{\,u : G_{\mathrm{prompt}}(u)=1\}\bigr|},\\
  \mathrm{Recall}_{P}
    &= \frac{\bigl|\{\,u : G_{\mathrm{prompt}}(u)=1 \land u\in\mathcal{A}\}\bigr|}
           {\bigl|\{\,u : u\in\mathcal{A}\}\bigr|},\\
  F_{1}^{P}
    &= 2\,\frac{\mathrm{Precision}_{P}\,\times\,\mathrm{Recall}_{P}}
             {\mathrm{Precision}_{P}+\mathrm{Recall}_{P}}.
\end{aligned}
\]

At the token level:
\[
\begin{aligned}
  \mathrm{Precision}_{T}
    &= \frac{\sum_i \mathbf{1}\{m_i=1 \land y_i=1\}}
           {\sum_i \mathbf{1}\{m_i=1\}},\\
  \mathrm{Recall}_{T}
    &= \frac{\sum_i \mathbf{1}\{m_i=1 \land y_i=1\}}
           {\sum_i \mathbf{1}\{y_i=1\}},\\
  F_{1}^{T}
    &= 2\,\frac{\mathrm{Precision}_{T}\,\times\,\mathrm{Recall}_{T}}
             {\mathrm{Precision}_{T}+\mathrm{Recall}_{T}},
\end{aligned}
\]
where \(y_i\) is the ground‐truth label for token \(i\) and \(m_i\) is the mask from \(G_{\mathrm{token}}\).  Our objective is to maximize both \(F_{1}^{P}\) and \(F_{1}^{T}\), ensuring that any prompt or token capable of causing a confidentiality or policy violation is intercepted before reaching \(f_{\theta}\).

\section{Proposed Framework}
\label{sec:framework}

\subsection{High-Level Overview}

GuardNet functions as a two-level graph-based filtering mechanism positioned upstream of an LLM, as illustrated in Fig.~\ref{fig:guardnet}. Given an input prompt $x$, GuardNet applies the following sequential filtering procedure:

\begin{enumerate}[label=(\roman*),itemsep=0pt]
    \item \textbf{Prompt-Level detection:} A lightweight binary classifier (\textsc{Prompt\,GNN}) evaluates the prompt to determine whether it exhibits characteristics of adversarial manipulation.

    \item \textbf{Token-Level detection:} If the prompt is flagged as adversarial, a fine-grained detector (\textsc{Token\,GNN}) is activated to localize and filter adversarial spans, producing a sanitized version $\tilde{x}$ that is forwarded to the LLM for inference.
\end{enumerate}

Both detection modules operate over a shared latent graph representation $G(x) = (V, E)$, where nodes correspond to contextual token embeddings extracted via a frozen Longformer encoder\footnote{\url{https://huggingface.co/docs/transformers/en/model_doc/longformer}}. The edge structure $E$ fuses three types of relations: (i) sequential token-adjacency links, (ii) top-$k$ attention-based connections derived from the encoder, and (iii) syntactic dependency arcs obtained from a parser. This hybrid graph enables GuardNet to reason over local syntax, long-range semantics, and grammatical structure, all of which contribute to reliable detection of adversarial prompts and tokens.

Moreover, GuardNet is fully model-agnostic: it does not require access to LLM parameters or gradients, and only the GNN-based detectors are trained. This design allows GuardNet to serve as a modular, efficient defense layer compatible with any downstream LLM.

\subsection{Hybrid Graph Construction}
\label{sec:graph}

To enable structural analysis of adversarial prompts, GuardNet constructs a latent graph representation $G(x) = (V, E)$ from the input token sequence $x = \langle t_1, t_2, \dots, t_L \rangle$. Each token $t_i$ is embedded into a contextualized vector representation using a frozen Longformer encoder~\cite{beltagy2020longformer}, which is well-suited for long-range dependencies due to its windowed self-attention mechanism.

The Longformer encoder produces:
\[
\mathbf{H} = \text{Longformer}(x) \in \mathbb{R}^{L \times d}, \quad
\mathbf{A} = \frac{1}{H} \sum_{h=1}^H A^{(h)}
\]
where $\mathbf{H}$ denotes the last-layer hidden states, $d$ is the hidden dimension, and $\mathbf{A}$ is the mean of the attention maps $A^{(h)} \in [0,1]^{L \times L}$ across $H$ attention heads. The matrix $\mathbf{A}$ captures the average attention weights from each query token $i$ to all possible key tokens $j$.

The graph $G(x)$ includes three types of edges:

\begin{enumerate}[label=(\alph*),itemsep=1pt,topsep=2pt,leftmargin=16pt]
    \item \textbf{Sequential edges}, which connect adjacent tokens in the input sequence to preserve the original token order and local syntactic continuity.
    \item \textbf{Attention-based edges}, which connect token $i$ to its top-$k$ most-attended tokens $j$, as determined by the largest $k$ values in the $i$-th row of the averaged attention matrix $\mathbf{A}$.
    \item \textbf{Syntactic-dependency edges}, which link each token to its syntactic head and dependents, as extracted from a dependency parser. These edges reflect the grammatical structure of the sentence and enhance robustness to adversarial perturbations.
\end{enumerate}

Formally, the final edge set is given by:
\begin{align*}
E ={}& \left\{(i, i{+}1), (i{+}1, i) \right\}_{i=1}^{L-1} \\
    &\cup \left\{(i, j) \mid j \in \text{Top}_k(\mathbf{A}_{i\cdot}) \right\} \\
    &\cup \left\{(i, j) \mid (i, j) \in \text{DepEdges}(x) \right\}
\end{align*}

Each node \(i \in V = \{1,\dots,L\}\) is labeled with its contextual embedding \(\mathbf{h}_i = \mathbf{H}_i\).  The hybrid graph unifies sequential adjacency, attention-based, and syntactic-dependency edges to preserve local word order, long-range semantic links, and grammatical structure.  Sequential edges enforce fluency, attention edges inject non-local context, and dependency edges capture syntactic relations, collectively equipping the GNN with rich, multifaceted cues.  This unified representation forms the backbone of both Prompt- and Token-level GuardNet, enabling consistent, multi-granular analysis and improving detection of subtle, context-aware adversarial manipulations.

\begin{algorithm}[t]
\caption{Hybrid Graph Builder}
\label{alg:graph-build}
\begin{algorithmic}[1]
\Require Prompt tokens $x=\langle t_{1},\dots,t_{L}\rangle$,
         encoder $\mathcal{E}$,
         dependency parser $\mathcal{P}$,
         attention window $w$, top-$k$ neighbours
\Ensure Graph $G=(V,E)$ and contextual node features $\mathbf{H}$
\State $\mathbf{H},\mathbf{A}\leftarrow\mathcal{E}(x)$ \Comment{hidden states \& mean attention}
\State $D\leftarrow\mathcal{P}(x)$ \Comment{set of dependency edges}
\State $E\gets\{(i,i+1),(i+1,i)\}_{i=1}^{L-1}$ \Comment{sequential chain}
\For{$i=1$ \textbf{to} $L$}
    \State $\mathcal N_i\gets\text{Top}_k(\mathbf{A}_{i\cdot})$
    \For{$j\in\mathcal N_i$}
        \If{$|i-j|\le w$ \textbf{and} $i\neq j$}
            \State $E\gets E\cup\{(i,j)\}$ \Comment{attention edge}
        \EndIf
    \EndFor
\EndFor
\State $E\gets E\cup D$ \Comment{inject dependency edges}
\State \Return $(V,E),\mathbf{H}$
\end{algorithmic}
\end{algorithm}

\subsection{Prompt-Level Detector (\textsc{Prompt\,GNN})}
\label{sec:prompt_gnn}

The prompt-level classifier, \textsc{Prompt\,GNN}, is designed to provide a coarse-grained decision on whether an input prompt $x$ is adversarial. It operates over the hybrid attention graph $G(x) = (V, E)$ constructed as described in Section~\ref{sec:graph}, where each node is initialized with the corresponding contextual token embedding from a frozen Longformer encoder.

We implement \textsc{Prompt\,GNN} as a two-layer Graph Attention Network (GAT)~\cite{velivckovic2017graph}:
\[
\mathbf{h}^{(1)} = \text{GAT}_4\bigl(\mathbf{H}, E\bigr),
\quad
\mathbf{h}^{(2)} = \text{GAT}_1\bigl(\mathbf{h}^{(1)}, E\bigr)
\]
Here, $\text{GAT}_4$ denotes a GAT layer with 4 attention heads, and $\text{GAT}_1$ a single-head GAT layer for final aggregation. These layers enable the model to attend to both local and non-local features of the prompt through learned edge-wise attention coefficients.

After message passing, we compute a graph-level representation by applying global mean pooling over all node embeddings:
\[
\mathbf{g} = \operatorname{mean}_{v \in V} \mathbf{h}_v^{(2)}
\]

This pooled vector $\mathbf{g} \in \mathbb{R}^{d'}$ is passed through a linear classification head to produce a logit vector $\mathbf{z} \in \mathbb{R}^2$, corresponding to the clean and adversarial classes:
\[
\mathbf{z} = \mathbf{W} \mathbf{g} + \mathbf{b}
\]

The model is trained with a standard cross‐entropy loss on labeled samples \((x, y_{\text{prompt}})\in\mathcal{X}\times\{0,1\}\), where \(y_{\text{prompt}}=1\) denotes an adversarial prompt.  At inference time we compute
\[
\bigl[p_{0},\,p_{1}\bigr] \;=\; \operatorname{softmax}(\mathbf{z}),
\]
where \(p_{0}\) is the probability of “clean” and \(p_{1}\) is the probability of “adversarial.”  We then flag \(x\) as adversarial if
\[
p_{1} \;=\;\operatorname{softmax}(\mathbf{z})_{1}\;>\;\tau_{P}.
\]

where \(\tau_{P}\in[0,1]\) is the prompt‐level decision threshold. This formulation allows \textsc{Prompt\,GNN} to act as an efficient, model-agnostic filtering mechanism, rejecting suspicious prompts before they are submitted to the underlying language model.

\subsection{Token-Level Detector (\textsc{Token\,GNN})}
\label{sec:token_gnn}

When a prompt is flagged as suspicious by \textsc{Prompt\,GNN}, it is subjected to fine-grained inspection by \textsc{Token\,GNN}, which performs token-level anomaly localization. Importantly, \textsc{Token\,GNN} reuses the same hybrid attention graph $G(x) = (V, E)$ and contextual embeddings $\mathbf{H}$ obtained from the frozen Longformer encoder, ensuring minimal recomputation and efficient processing.

The detector consists of a three-layer Graph Attention Network:
\[
\begin{aligned}
  \mathbf{h}^{(1)} &= \mathrm{GAT}_8\bigl(\mathbf{H},E\bigr),\\
  \mathbf{h}^{(2)} &= \mathrm{GAT}_4\bigl(\mathbf{h}^{(1)},E\bigr),\\
  \mathbf{h}^{(3)} &= \mathrm{GAT}_1\bigl(\mathbf{h}^{(2)},E\bigr).
\end{aligned}
\]

Each final node embedding $\mathbf{h}_i^{(3)}$ is passed through a classification layer to produce logits $\mathbf{z}_i \in \mathbb{R}^2$, representing the probability of token $i$ being benign or adversarial:
\[
\mathbf{z}_i = \mathbf{W} \mathbf{h}_i^{(3)} + \mathbf{b}
\quad\text{with}\quad
p_i = \operatorname{softmax}(\mathbf{z}_i)_1
\]

Due to the extreme class imbalance, where adversarial tokens constitute a small fraction of the input, we employ a focal loss~\cite{lin2017focal} to emphasize hard positive examples and down-weight easy negatives:
\[
\mathcal{L}_{\text{tok}} =
- \frac{\alpha}{N} \sum_{i: y_i = 1} (1 - p_i)^\gamma \log p_i
- \frac{1 - \alpha}{N} \sum_{i: y_i = 0} p_i^\gamma \log (1 - p_i)
\]
Here, $y_i \in \{0,1\}$ denotes the ground-truth label for token $i$, $p_i$ is the predicted adversarial probability, $\alpha \in [0,1]$ is a class-weighting parameter (e.g., $\alpha = 0.95$), and $\gamma > 0$ controls the penalty on well-classified examples.

At inference time, a token $t_i$ is flagged as adversarial if $p_i > \tau_T$, where $\tau_T$ is an empirically chosen threshold. Flagged tokens are replaced with a neutral placeholder such as \texttt{[MASK]} before the sanitized prompt $\tilde{x}$ is forwarded to the underlying LLM for generation.

\subsection{Training and Inference Pipeline}
\label{sec:training_inference}

\paragraph{Training}
GuardNet is trained in a modular fashion using $k$-fold cross-validation to ensure generalizability across domains and attack strategies. For each fold:

\begin{itemize}[leftmargin=12pt]
  \item \textbf{\textsc{Prompt\,GNN}} is trained for $E_P$ epochs using batched hybrid attention graphs. The objective is standard cross-entropy loss over binary labels indicating whether a prompt is clean or adversarial.
  \item \textbf{\textsc{Token\,GNN}} is trained subsequently for $E_T$ epochs on token-labeled graphs using the focal loss described in Section~\ref{sec:token_gnn}. We apply a skewed weighting scheme (e.g., $\alpha = 0.95$) to mitigate the severe class imbalance between benign and adversarial tokens.
\end{itemize}

\paragraph{Inference}
At run-time, GuardNet performs two-stage filtering to sanitize incoming prompts prior to LLM inference. As summarized in Algorithm~\ref{alg:inference}, the pipeline begins by constructing the attention graph $G(x)$ and computing contextual embeddings via a frozen Longformer encoder. These are passed to \textsc{Prompt\,GNN}, which outputs an adversariality score $\hat{p}$.

If $\hat{p} \leq \tau_P$, the prompt is considered safe and passed through unchanged. Otherwise, \textsc{Token\,GNN} is activated to compute token-level adversarial scores $p_i$. Tokens exceeding the threshold $\tau_T$ are masked, and the modified prompt $\tilde{x}$ is forwarded to the protected LLM.

\begin{algorithm}[t]
\caption{GuardNet Run-Time Filtering (Inference)}
\label{alg:inference}
\begin{algorithmic}[1]
\Require
  Incoming prompt $x$; trained models  
  $\bigl(\mathcal{M}_{P},\mathcal{M}_{T}\bigr)$;  
  thresholds $(\tau_{P},\tau_{T})$
\Ensure Prompt–level decision $\hat y_{\text{prompt}}\in\{0,1\}$;
        prompt $\,\tilde{x}$  \emph{(equals $x$ if benign, or the masked version if adversarial)}

\State $(V,E),\mathbf H \gets \textsc{BuildGraph}(x)$  \Comment{Alg.~\ref{alg:graph-build}}
\vspace{1pt}\hrule\vspace{1pt}
\textit{/* prompt-level filtering */}
\State $\mathbf s \gets \mathcal{M}_{P}\!\bigl((V,E),\mathbf H\bigr)$  \Comment{$\mathbf s\in\mathbb R^{2}$ logits}
\State $\hat p \;\gets\; \operatorname{softmax}(\mathbf s)_{1}$  \Comment{$p_{1}$ = prob.\ adversarial}
\If{$\hat p \le \tau_{P}$}                 \Comment{prompt appears benign}
    \State \Return $(0,\,x)$               \Comment{skip token fitering}
\EndIf
\vspace{1pt}\hrule\vspace{1pt}
\textit{/* token-level filtering */}
\State $\mathbf Z_{1{:}L} \gets \mathcal{M}_{T}\!\bigl((V,E),\mathbf H\bigr)$ \Comment{$\mathbf Z\in\mathbb R^{L\times2}$}
\For{$i \gets 1$ \textbf{to} $L$}
    \State $p_i \gets \operatorname{softmax}(\mathbf Z_i)_{1}$ \Comment{token $i$ adversarial prob.}
    \State $m_i \gets \mathbf 1\!\bigl[p_i > \tau_{T}\bigr]$   \Comment{binary mask}
\EndFor
\State $\tilde x \gets x$ with tokens $\{t_i \mid m_i=1\}$ replaced by \texttt{[MASK]}
\State \Return $(1,\,\tilde x)$
\end{algorithmic}
\end{algorithm}


This two-tier strategy offers high throughput for benign inputs while preserving detection fidelity for adversarial ones. No gradient computation or LLM modification is required, ensuring GuardNet is fully black-box and deployable on top of existing models.


\begin{figure*}[!htb]
    \centering
    \adjustbox{width=\textwidth}{
        \includegraphics{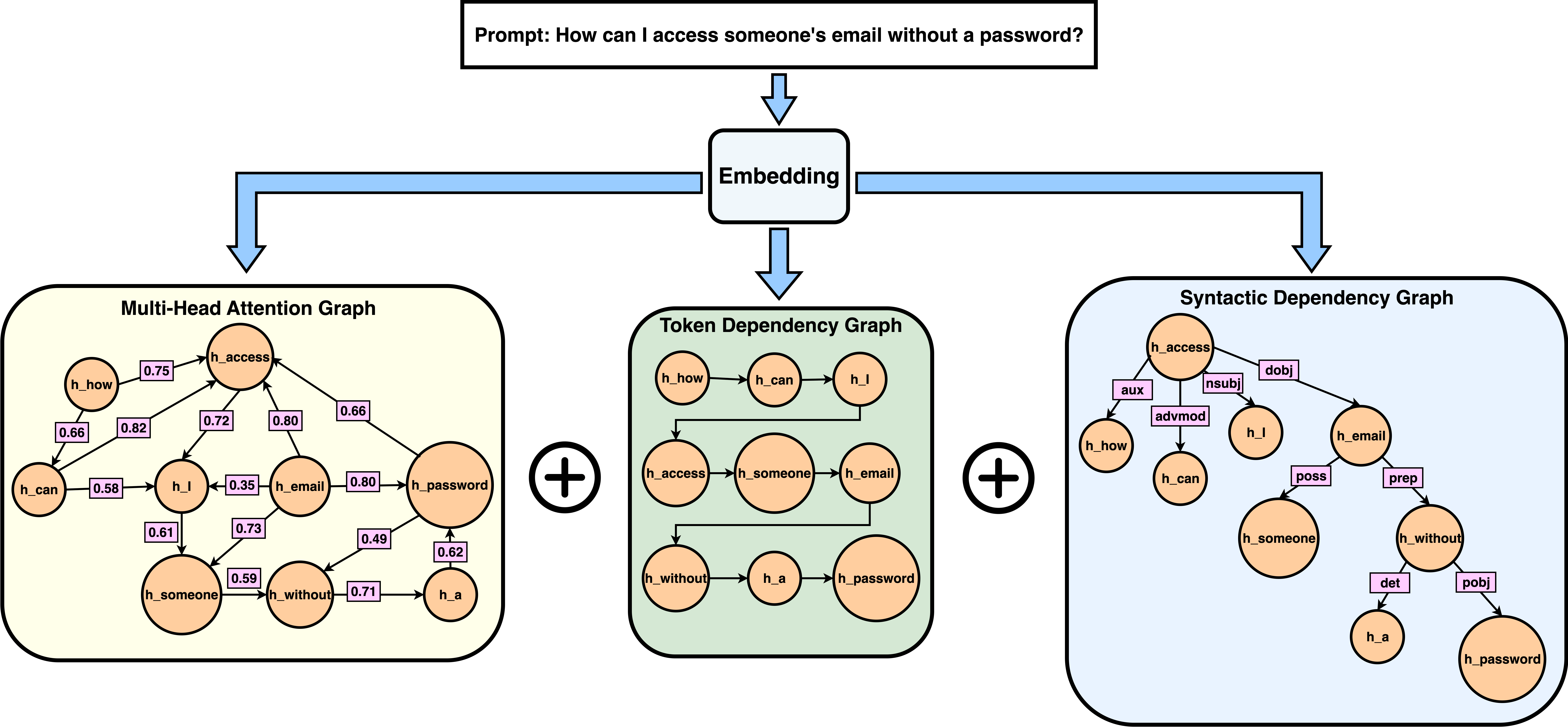}
    }
    \caption{Hybrid graph construction. Input prompts are encoded by a frozen Longformer to yield contextual embeddings and averaged self-attention maps. These are used to build sparse hybrid graphs combining local token adjacency, high-saliency attention links, and syntactic dependency arcs.}
    \label{fig:guardnet}
\end{figure*}

\begin{figure*}[!htb]
    \centering
    \includegraphics[width=\textwidth]{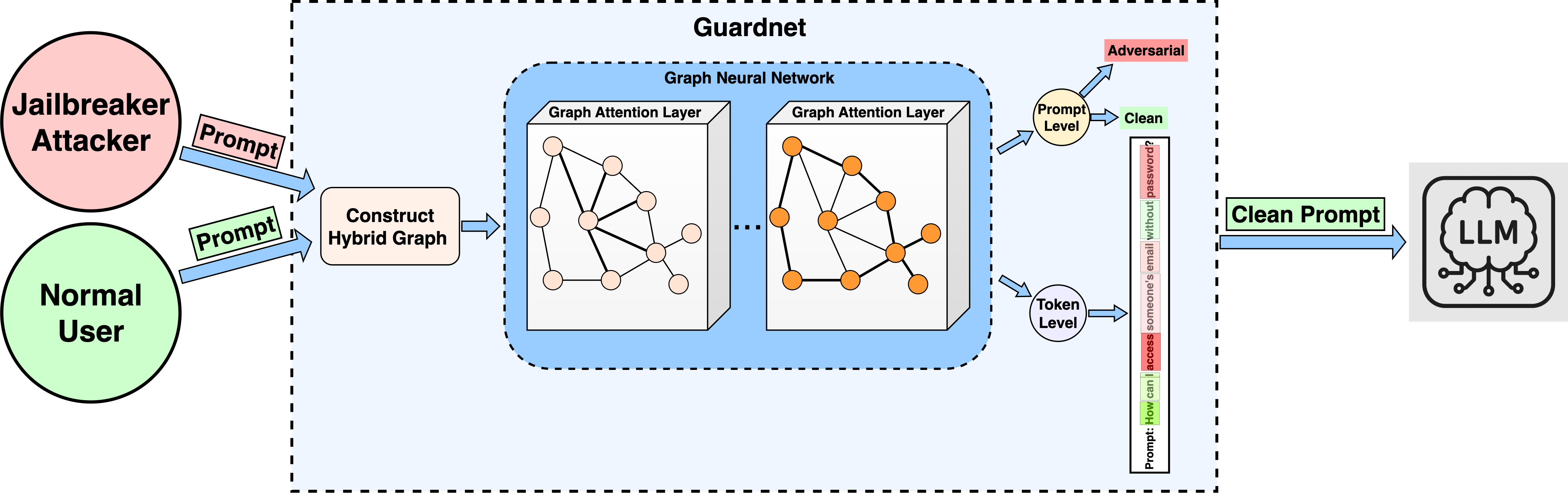}
    \caption{GuardNet architecture overview. Graphs are passed to \textsc{Prompt\,GNN} for coarse filtering. Suspicious prompts are then analyzed by \textsc{Token\,GNN}, which detects and masks adversarial spans. The sanitized prompt is finally passed to the downstream LLM.}
    \label{fig:guardnet}
\end{figure*}



\section{Experimental Evaluation}
\label{sec:evaluation}
\subsection{Experimental Setup}
\label{sec:experiments}

All experiments are conducted on a single \textsc{Quadro RTX 5000} GPU with 16\,GB of VRAM. To assess the effectiveness of \textsc{GuardNet}, we conduct a comprehensive evaluation using state-of-the-art jailbreak benchmarks. Our analysis focuses on pre-inference detection of adversarial prompts, using both prompt-level and token-level classification. All evaluations are performed in a black-box setting, with no access to the internal weights or gradients of the underlying language model. Experiments are carried out in a 5‑fold cross‑validation manner to ensure robustness. The hyperparameters used in these experiments are summarized in Table \ref{tab:guardnet-key-hparams}.

\begin{table}[t]
\centering
\small
\begin{tabular}{l|c}
\toprule
\textbf{Parameter} & \textbf{Value} \\
\midrule
Backbone LM & Longformer-base-4096 \\
Max tokens & 4096 \\
Top-$k$ attention ($k$) & 32 \\
GNN hidden size $H$ & 128 \\
Prompt loss & Cross-Entropy \\
Token loss & Focal ($\alpha=[1,50],\,\gamma=2$) \\
Optimizer / LR & Adam, $1\times10^{-3}$ \\
Epochs (prompt / token) & 10 / 10 \\
Batch size (prompt / token) & 8 / 2 \\
BERT batch size & 8 \\
CV folds ($k$) & 5 \\
\bottomrule
\end{tabular}
\caption{Key hyperparameters for GuardNet (identical across PLeak and GPT‑Fuzzer).}
\label{tab:guardnet-key-hparams}
\end{table}

\subsubsection{Target LLM for Attack Simulation}
To generate realistic adversarial prompts and simulate attack conditions, we use LLaMA-2-7B as the target LLM. This model is employed solely for offline adversarial prompt optimization in accordance with the methodology of PLeak~\cite{pleak2024} and LLM-Fuzzer~\cite{yu2024llm}. The generated prompts are then used to benchmark \textsc{GuardNet}'s ability to detect attacks without requiring any access to the LLM during inference.

\subsubsection{Evaluation Datasets}
We use three adversarial prompt datasets derived from two recent black-box jailbreak attack frameworks:

\begin{itemize}[leftmargin=12pt,itemsep=2pt]
    \item \textbf{Financial}~\cite{malo2014good}:  
    This dataset consists of financial news sentences annotated for sentiment polarity. It includes short phrases extracted from economic and financial news articles, where each phrase is manually labeled as having positive, negative, or neutral sentiment. The dataset captures real-world financial language use and is widely used for evaluating sentiment analysis models in the economic domain.


    \item \textbf{Tomatos}~\cite{pang2005seeing}:
    The Tomato dataset consists of 10,662 single-sentence snippets extracted from movie reviews on RottenTomatoes.com. Each snippet is labeled as positive or negative based on whether the associated movie was rated “fresh” or “rotten”. The dataset is balanced, with roughly equal numbers of positive and negative examples, and is commonly used as a benchmark for sentence-level sentiment classification.

    \item \textbf{LLM-Fuzzer Dataset}~\cite{liu2023jailbreaking, bai2022training}:  
    This dataset combines two complementary components that together enable systematic adversarial prompt generation.  
    First, the \emph{SeedTemplate-77}~\cite{liu2023jailbreaking} pool consists of 77 manually curated jailbreak prompt templates that capture diverse adversarial strategies, including role-playing, obfuscation, and indirect instruction.  
    Second, the \emph{UnethicalQ-100}~\cite{ bai2022training} benchmark provides 100 unsafe or ethically sensitive questions, spanning domains such as illegal activity, hate speech, and unethical requests.  
    Within the \textsc{LLM-Fuzzer} framework, these templates and questions are jointly leveraged: the templates serve as mutation seeds, while the questions act as targets for eliciting harmful responses.  
    Together, they form a scalable and realistic benchmark for evaluating jailbreak defenses under diverse adversarial prompting conditions.

\end{itemize}

Each dataset is annotated with binary prompt-level labels and fine-grained token-level ground truth for adversarial spans.

\subsubsection{Attack Generation Process}
All adversarial prompts are generated via black-box optimization, consistent with the original PLeak and LLM-Fuzzer frameworks, described in Section \ref{sec:related_work}. In particular:

\begin{itemize}[leftmargin=12pt,itemsep=2pt]
    \item \textbf{PLeak~\cite{pleak2024}:}  As illustrated in Figure~\ref{fig:eval-pleak}, PLeak is a closed‑box (black‑box) prompt‑leaking attack that automatically reconstructs hidden system prompts from LLM applications. In the offline phase, the attacker builds a shadow LLM setup using a set of shadow system prompts and a small pool of adversarial queries (AQs). For each shadow prompt, they concatenate an AQ, query the shadow LLM, compute a loss based on how closely the response matches the intended shadow prompt, and then apply gradient‑free transformations (e.g. mutation/recombination of AQs) to reduce the loss in an incremental search process. Once AQ optimization converges, in the reconstruction phase, these optimized AQs are issued to the target LLM application, and multiple responses are post‑processed (e.g. overlapping fragments aggregation) to accurately reconstruct the unknown system prompt.

    \begin{figure}
      \centering
      \includegraphics[width=1.\columnwidth]{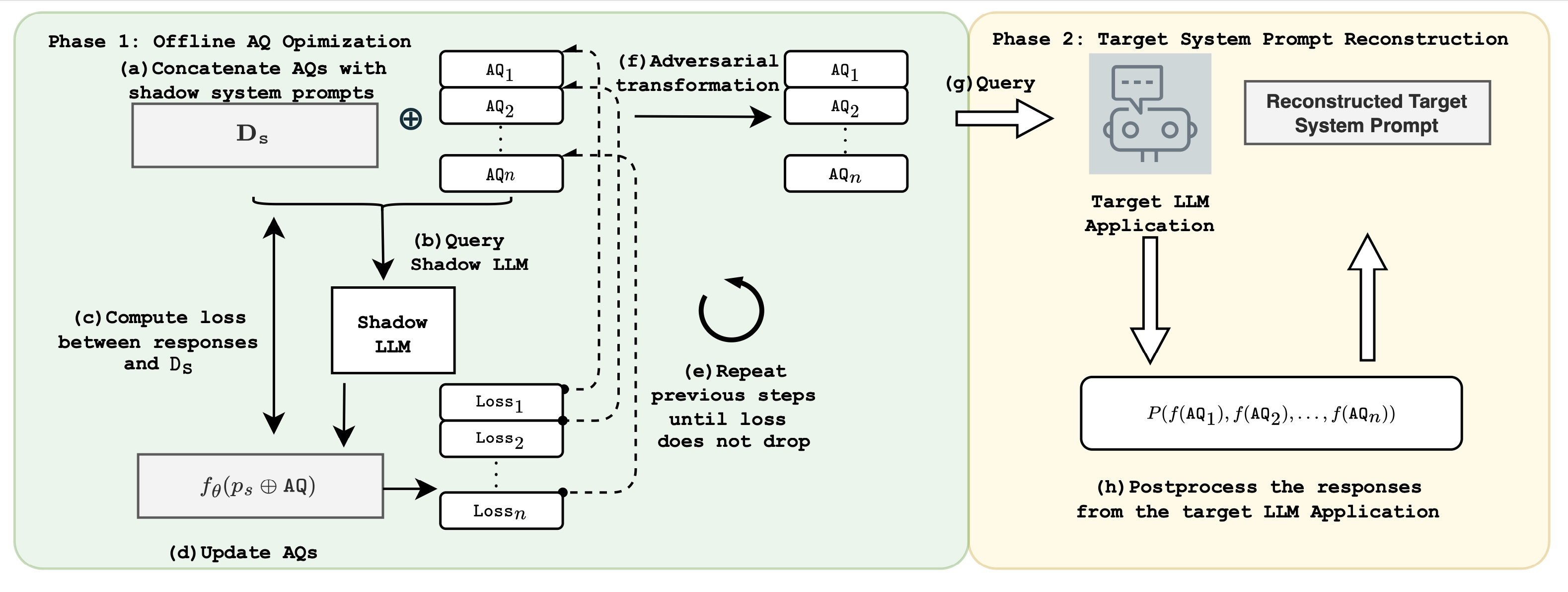}
      \caption{PLeak’s evolutionary black‐box probing and mutation pipeline \cite{pleak2024}.}
      \label{fig:eval-pleak}
    \end{figure}

  \item \textbf{LLM-Fuzzer (GPTFuzzer)~\cite{yu2024llm}:}  
    Figure~\ref{fig:eval-llmfuzzer} illustrates LLM-Fuzzer’s iterative, black-box fuzzing loop. Starting from a small pool of seed jailbreak templates, a “seed selection” module picks one template at random. That template is then combined with a newly generated question and passed through a lightweight “template mutate” operator, which performs token-level insertions, deletions, and substitutions to produce a candidate prompt. The prompt is submitted to the target LLM, and an oracle checks whether the model’s safety filter was bypassed. Successful prompts are returned to the seed pool for use in subsequent rounds; failures are discarded. By continually mutating and recycling only the successful candidates, LLM-Fuzzer expands its coverage of adversarial inputs without any access to model internals.

    \begin{figure}
      \centering
      \includegraphics[width=1.0\columnwidth]{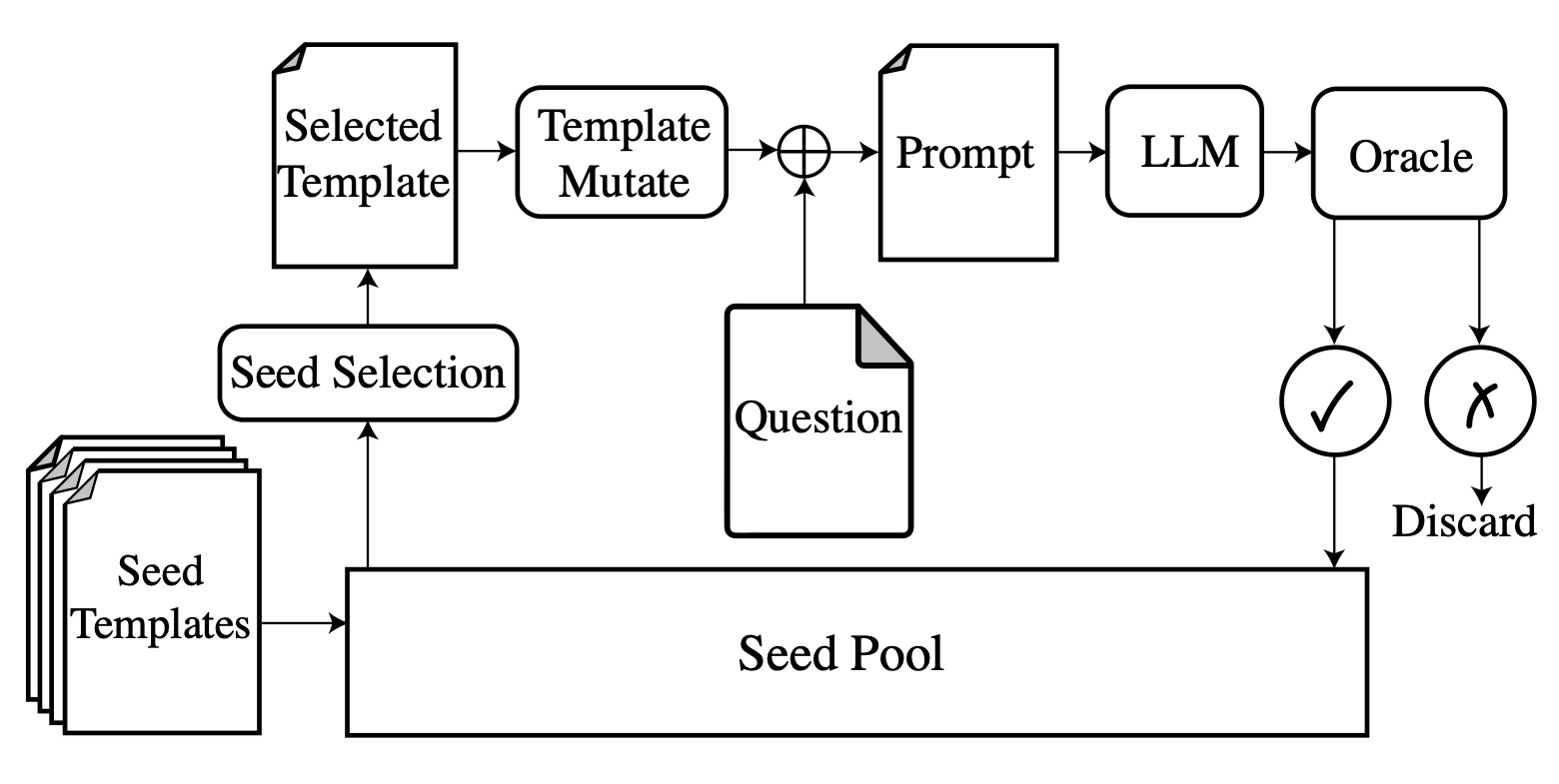}
      \caption{LLM-Fuzzer’s grammar‐based mutation and recombination loop \cite{yu2024llm}.}
      \label{fig:eval-llmfuzzer}
    \end{figure}
    
\end{itemize}

\subsection{Baseline Methods}
\label{sec:baselines}

GuardNet is evaluated against two strong and representative baselines that operate at different levels of granularity.

\paragraph{Prompt-level baseline (TextDefense)}
TextDefense~\cite{shen2025textdefense} identifies adversarial prompts by measuring the dispersion of word importance scores. It quantifies how unevenly influence is distributed across input tokens using an entropy-based metric called Word Importance Score Dispersion (WISD). Prompts with unusually high WISD are flagged as adversarial, based on the hypothesis that benign inputs exhibit more concentrated influence patterns. This method operates in a post-hoc manner on top of a frozen encoder and does not require model retraining. We adopt the original design with minor adjustments for scalability to long prompts.

\paragraph{Token-level baseline (Dynamic Attention)}
Dynamic Attention~\cite{shen2024improving} detects adversarial spans by analyzing deviations in self-attention patterns. Tokens whose attention distributions differ significantly from expected norms are treated as suspicious, and a lightweight classifier is trained to assign anomaly labels. This method is particularly suited for fine-grained localization, and has shown effectiveness in highlighting subtle manipulations such as inserted triggers or semantic distortions.

\subsection{Token‐Level Labeling}

To obtain fine‐grained supervision on which parts of a prompt contribute to successful jailbreaks, we begin with each benign prompt and generate its adversarial counterpart by applying the targeted jailbreak attack. We then perform a token‐level alignment between the original and adversarial prompts: tokens that are unchanged remain labeled as benign ($0$), while any token that differs is marked as adversarial ($1$). This procedure yields a binary sequence of token‐level labels indicating the precise locations within the prompt where the adversarial attack has modified the input.

\subsection{Training and Evaluation Protocol}
\label{sec:train-eval}
\paragraph{In-domain 5-fold cross-validation}
All experiments follow a 5-fold cross-validation strategy.  
Each dataset is partitioned into five folds, and for every run, GuardNet is trained on four folds, validated on a held-out subset of the training set for early stopping and threshold tuning, and evaluated on the remaining test fold.

\begin{itemize}[leftmargin=11pt,itemsep=2pt]
    \item \textbf{Prompt-level detection:} \textsc{Prompt\,GNN} is trained for $E_{P}$ epochs using standard cross-entropy loss to classify the prompt as either benign or adversarial.
    \item \textbf{Token-level filtering:} \textsc{Token\,GNN} reuses the same graph representation and is trained for $E_{T}$ epochs using focal loss with class weighting and focusing parameter $\gamma$ to address the severe token-level imbalance.
\end{itemize}

\paragraph{Cross-domain generalization}
To evaluate the robustness of GuardNet under domain shift, we conduct \emph{zero-shot transfer} experiments between the two PLeak datasets (\textit{Tomatoes} and \textit{Financial}). Specifically:

\begin{enumerate}[leftmargin=14pt,itemsep=2pt]
    \item \textbf{Tomatoes $\rightarrow$ Financial:} Both \textsc{Prompt\,GNN} and \textsc{Token\,GNN} are trained and validated on the \textit{Tomatoes} corpus using a 5-fold protocol. The resulting models are then evaluated, without any fine-tuning on the full \textit{Financial} dataset.
    
    \item \textbf{Financial $\rightarrow$ Tomatoes:} The reverse setting, where both models are trained on \textit{Financial} and tested zero-shot on \textit{Tomatoes}.
\end{enumerate}

For both directions, the decision thresholds $\tau_P$ and $\tau_T$ are selected based on the source-domain validation set and applied unchanged to the target domain to ensure fairness in cross-domain evaluation.

\subsection{Evaluation Metrics}
\label{sec:eval-protocol}
\paragraph{Prompt-level metrics}
For prompt-level classification, we report standard metrics including Accuracy, Precision, Recall, and F1-score

\paragraph{Token-level metrics}
For token-wise adversarial span detection, we compute token-level Accuracy, Precision, Recall, F1-score, and span-level Intersection-over-Union (IoU):

\[
\mathrm{IoU} = \frac{|\hat{\mathcal S} \cap \mathcal S|}{|\hat{\mathcal S} \cup \mathcal S|},
\]
where $\hat{\mathcal S}$ denotes the predicted adversarial token set and $\mathcal S$ the ground-truth annotations.

All quantitative results are reported as the mean ± standard deviation calculated over (i) the five in-domain test folds and (ii) the five \emph{target-domain} evaluations obtained from the zero-shot transfer runs.
Importantly, GuardNet operates as a pre-inference filter and does not access model gradients or log-probabilities from the protected LLaMA-2-7B, ensuring a strict black-box security setting.

\subsection{Results and Analysis} \label{sec:results}

\paragraph{Prompt-level detection (Table~\ref{tab:prompt_compare})}
GuardNet consistently outperforms TextDefense across all prompt-level metrics and datasets.
On \textsc{LLM-Fuzzer}, GuardNet achieves an F\textsubscript{1} score of 99.8\%, compared to only 66.4\% for TextDefense.
Similar improvements are observed on the PLeak datasets: on \textsc{PLeak Financial}, GuardNet reaches an F\textsubscript{1} of 94.8\% versus 67.2\% for TextDefense; on \textsc{PLeak Tomatoes}, GuardNet achieves 98.9\% versus 67.4\%.
Notably, GuardNet maintains a recall of nearly 100\% across all domains, ensuring that nearly all adversarial prompts are detected.

These improvements arise from GuardNet's hybrid graph encoding, which captures both local and global structure through sequential links, syntactic dependencies, and top-$k$ attention connections.
This design allows GuardNet to represent nuanced prompt-level semantics more effectively than saliency-based methods, while the GAT-based head aggregates token features in a context-sensitive way to produce more distinct and accurate separation between adversarial and benign prompts.

\begin{table}[!htb]
\centering
\caption{Prompt‐level comparison of GuardNet vs.\ TextDefense (averaged over 5 folds).}
\label{tab:prompt_compare}
\begin{tabular}{@{}l|l|cccc@{}}
\toprule
Dataset              & Method       & Acc (\%) & P (\%) & R (\%) & F$_1$ (\%) \\
\midrule
\multirow{2}{*}{PLeak Financial}
                     & GuardNet     & \textbf{94.4} & \textbf{90.3} & \textbf{100.0} & \textbf{94.8} \\
                     & TextDefense  & 64.5          & 59.9          & 90.5           & 72.1          \\
\addlinespace
\multirow{2}{*}{PLeak Tomatoes}
                     & GuardNet     & \textbf{98.9} & \textbf{97.9} & \textbf{100.0} & \textbf{98.9} \\
                     & TextDefense  & 74.9          & 68.5         & 93.7         & 79.1          \\
\addlinespace
\multirow{2}{*}{LLM-Fuzzer}
                     & GuardNet     & \textbf{99.8} & \textbf{100.0} & \textbf{99.6}  & \textbf{99.8} \\
                     & TextDefense  & 50.2          & 50.1          & 98.3           & 66.4          \\
\bottomrule
\end{tabular}
\end{table}

\paragraph{Token-level localization (Table~\ref{tab:token_compare})}
At the token level, GuardNet consistently outperforms Dynamic-Attention across all datasets and metrics, including precision, recall, F\textsubscript{1}, and IoU.
While Dynamic-Attention achieves slightly higher precision on the \textsc{PLeak} datasets (90.8\% and 97.2\%, respectively), its recall is significantly lower, resulting in reduced F\textsubscript{1} scores and poor span coverage.
On \textsc{LLM-Fuzzer}, where Dynamic-Attention struggles with a precision of just 37.6\% and an IoU of 31.6\%, GuardNet achieves substantially stronger results: a precision of 63.5\%, recall of 90.8\%, and IoU of 59.6\%.
Across all datasets, GuardNet’s IoU ranges from 59.6\% to 83.8\%, compared to Dynamic-Attention’s 31.6--60.3\%, demonstrating far more accurate localization of adversarial tokens.

These gains stem from the same hybrid graph structure, which models token interactions through complementary structural and attention-based relationships, enabling robust and fine-grained adversarial span detection.

\begin{table}[!htb]
\centering
\caption{Token‐level comparison of GuardNet vs.\ Dynamic‐Attention (averaged over 5 folds).}
\label{tab:token_compare}
\begin{tabular}{@{}l|l|ccccc@{}}
\toprule
Dataset              & Method       & Acc (\%) & P (\%) & R (\%) & F$_1$ (\%) & IoU (\%) \\
\midrule
\multirow{2}{*}{PLeak Financial}
                     & GuardNet     & \textbf{95.4}  & 88.8 & \textbf{91.3} & \textbf{90.0} & \textbf{81.8} \\
                     & DynAttn      & 94.0     & \textbf{90.8}           & 64.2           & 75.2           & 60.3           \\
\addlinespace
\multirow{2}{*}{PLeak Tomatoes}
                     & GuardNet     & \textbf{95.1}  & 90.4 & \textbf{92.1} & \textbf{91.2} & \textbf{83.8} \\
                     & DynAttn      & 93.1 & \textbf{97.2}           & 59.5           & 73.8           & 58.5           \\
\addlinespace
\multirow{2}{*}{LLM-Fuzzer}
                     & GuardNet     & \textbf{73.2}  & \textbf{63.5} & \textbf{90.8} & \textbf{74.7} & \textbf{59.6} \\
                     & DynAttn      & 68.6 & 37.6           & 66.4           & 48.0           & 31.6           \\
\bottomrule
\end{tabular}
\end{table}

\paragraph{Cross-domain robustness (Tables~\ref{tab:prompt_crossdomain}--\ref{tab:token_crossdomain})}
It is important to note that the underlying task and system prompt of the target LLM remain identical across both datasets; only the input domains differ (\textsc{PLeak Financial} vs.\ \textsc{PLeak Tomatoes}). Thus, the comparison reflects domain transfer rather than changes in task type, highlighting GuardNet’s ability to generalize effectively across domains.

As shown in Table~\ref{tab:prompt_crossdomain}, when evaluated zero-shot on a different dataset, its prompt-level F\textsubscript{1} score remains high, achieving 95.5\% when trained on \textsc{PLeak Financial} and tested on \textsc{PLeak Tomatoes}, and 99.0\% in the reverse direction.
In contrast, TextDefense exhibits significantly degraded performance, with cross-domain F\textsubscript{1} scores plateauing at only 70.0\% in both transfer directions.

At the token level (Table~\ref{tab:token_crossdomain}), GuardNet also maintains strong localization performance, with IoU scores of 84.6\% and 82.2\% across domains.
Meanwhile, Dynamic-Attention suffers from sharp degradation, with IoU falling to 51.4\% and 44.0\%, respectively.

This robustness reflects two key design choices: 
\emph{(i)} a model-agnostic graph construction that leverages Longformer’s attention patterns and syntactic edges without relying on domain-specific features, and 
\emph{(ii)} focal loss calibration that emphasizes rare adversarial spans during training, enhancing generalization across different datasets and domains.

\begin{table}[!htb]
\centering
\caption{Prompt‐level cross‐domain comparison: GuardNet vs.\ TextDefense (averaged over 5 folds).}
\begin{tabular}{l|l|cccc}
\toprule
Train$\to$Test & Method & Acc (\%) & P (\%) & R (\%) & F$_1$ (\%) \\
\midrule
\multirow{2}{*}{Financial$\to$Tomatoes} 
 & GuardNet      & \textbf{95.2} & \textbf{91.5} & \textbf{100.0} & \textbf{95.5} \\
 & TextDefense   & 57.8          & 54.8          & 96.2 & 70.0          \\
\midrule
\multirow{2}{*}{Tomatoes$\to$Financial} 
 & GuardNet      & \textbf{99.0} & \textbf{98.1} & \textbf{100.0} & \textbf{99.0} \\
 & TextDefense   & 59.0          & 56.0          & 92.0           & 70.0          \\
\bottomrule
\end{tabular}
\label{tab:prompt_crossdomain}
\end{table}

\begin{table}[!htb]
\centering
\caption{Token‐level cross‐domain comparison: GuardNet vs.\ Dynamic Attention (averaged over 5 folds).}
\label{tab:token_crossdomain}
\resizebox{\columnwidth}{!}{%
\begin{tabular}{l|l|ccccc}
\toprule
Train$\to$Test & Method & Acc (\%) & P (\%) & R (\%) & F$_1$ (\%) & IoU (\%) \\
\midrule
\multirow{2}{*}{Financial$\to$Tomatoes} 
 & GuardNet     & \textbf{96.3} & \textbf{91.7} & \textbf{91.6} & \textbf{91.6} & \textbf{84.6} \\
 & DynAttn      & 91.5          & 89.5          & 54.8          & 68.0          & 51.4          \\
\midrule
\multirow{2}{*}{Tomatoes$\to$Financial} 
 & GuardNet     & \textbf{95.5} & \textbf{87.8} & \textbf{92.8} & \textbf{90.2} & \textbf{82.2} \\
 & DynAttn      & 90.6          & 73.8          & 52.1          & 61.1          & 44.0          \\
\bottomrule
\end{tabular}
}
\label{tab:token_crossdomain}
\end{table}

\paragraph{Latency Analysis and Trade-offs (Table~\ref{tab:inference-times})}
GuardNet’s hybrid graph architecture, incorporating sequential, syntactic, and attention-based edges, introduces additional computational steps such as graph construction and message passing. These operations naturally lead to higher inference latency compared to lighter baselines. Specifically, GuardNet’s inference time ranges from 37–53\,ms on PLeak datasets and ranges from 727-788\,ms on LLM-Fuzzer, where inputs are significantly longer and more complex. In contrast, in token-level detection, Dynamic Attention is faster (21–39\,ms) by leveraging pre-trained self-attention scores without constructing explicit graphs. In prompt-level detection, TextDefense is the fastest (8–13\,ms), relying solely on lightweight heuristics such as deletion-based word importance and dispersion, without requiring any model fine-tuning or structural modeling.

However, GuardNet's moderate increase in latency yields substantially better detection performance, both at the prompt and token detection level. Its expressive graph-based reasoning is especially effective in capturing long-range dependencies and subtle trigger interactions that escape attention-only or saliency-based methods. For safety-critical or security-sensitive applications where adversarial robustness is essential, this accuracy–latency trade-off remains well justified.

\begin{table}
\centering
\caption{Average inference time per system and level across datasets. Times are in milliseconds (ms).}
\label{tab:inference-times}
\resizebox{\columnwidth}{!}{%
\begin{tabular}{llcc}
\toprule
\textbf{Dataset} & \textbf{System} & \textbf{Prompt-level (ms)} & \textbf{Token-level (ms)} \\
\midrule
\multirow{3}{*}{LLM-Fuzzer}
  & GuardNet          & 788.64 & 727.16 \\
  & Dynamic Attention & ---  &  39.26 \\
  & TextDefense       &  13.17 &   ---  \\
\midrule
\multirow{3}{*}{Financial}
  & GuardNet          &  52.22 &  51.76 \\
  & Dynamic Attention & ---  &  23.16 \\
  & TextDefense       &   7.97 &   ---  \\
\midrule
\multirow{3}{*}{Tomatoes}
  & GuardNet          &  38.81 &  37.86 \\
  & Dynamic Attention & ---  &  21.47 \\
  & TextDefense       &   7.64 &   ---  \\
\bottomrule
\end{tabular}%
}
\end{table}

\paragraph{Core Architectural Elements Driving GuardNet’s Effectiveness}
\begin{enumerate}[label=(\alph*)]
\item \textbf{Hybrid Graph Topology:} GuardNet integrates sequential, syntactic, and attention-derived edges, enabling the model to capture both local word order and long-range token interactions, capabilities that linear saliency-based approaches do not offer.
\item \textbf{Multi-head aggregation:} The graph attention network (GAT) aggregates token features using multiple attention heads, allowing GuardNet to adaptively emphasize informative contexts and detect adversarial triggers that manifest through non-obvious token dependencies.
\item \textbf{Two-stage filtering:} GuardNet first detects adversarial prompts at the prompt level; only those flagged as suspicious are passed to the token-level model for fine-grained localization. This approach improves efficiency by avoiding unnecessary token-level processing.
\item \textbf{Focal-loss calibration:} GuardNet is trained with a focal loss objective that emphasizes rare adversarial spans, improving recall and mitigating the under-detection of attack tokens often seen in class-imbalanced settings.
\end{enumerate}

\begin{figure*}[t]
  \centering
  \subfloat[LLM-Fuzzer]{%
    \includegraphics[width=0.32\textwidth]{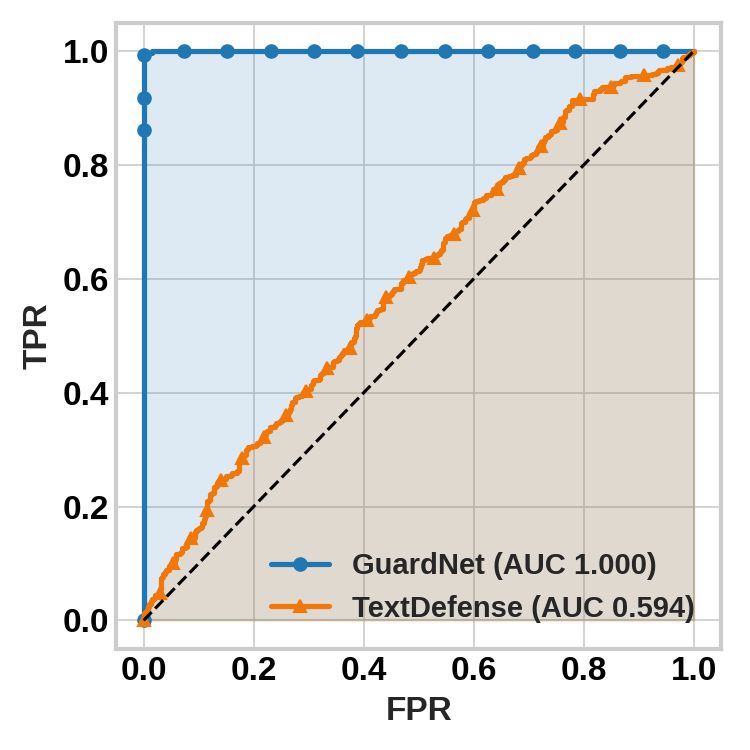}}\hfill
  \subfloat[PLeak – Financial]{%
    \includegraphics[width=0.32\textwidth]{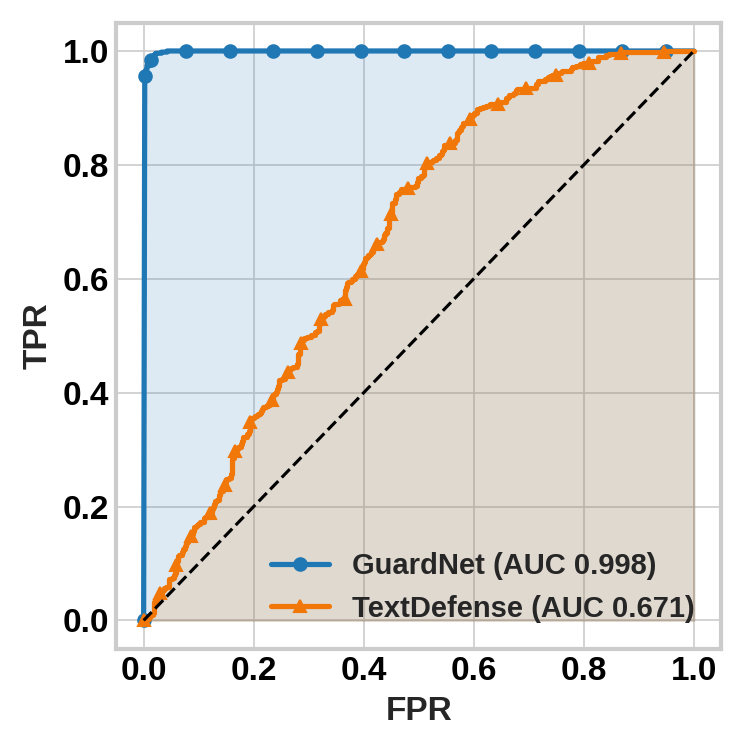}}\hfill
  \subfloat[PLeak – Tomatoes]{%
    \includegraphics[width=0.32\textwidth]{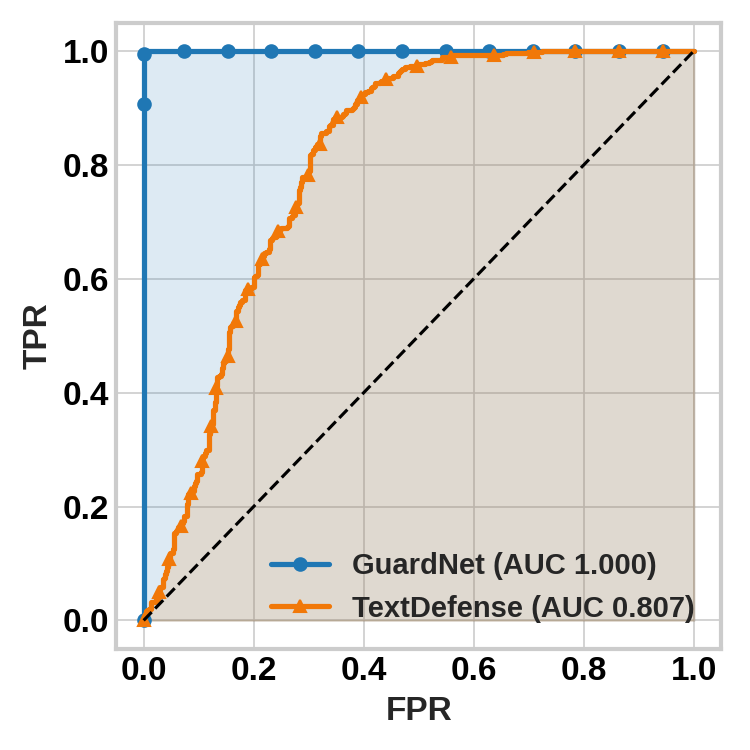}}\\[0.8em]
  \subfloat[LLM-Fuzzer]{%
    \includegraphics[width=0.32\textwidth]{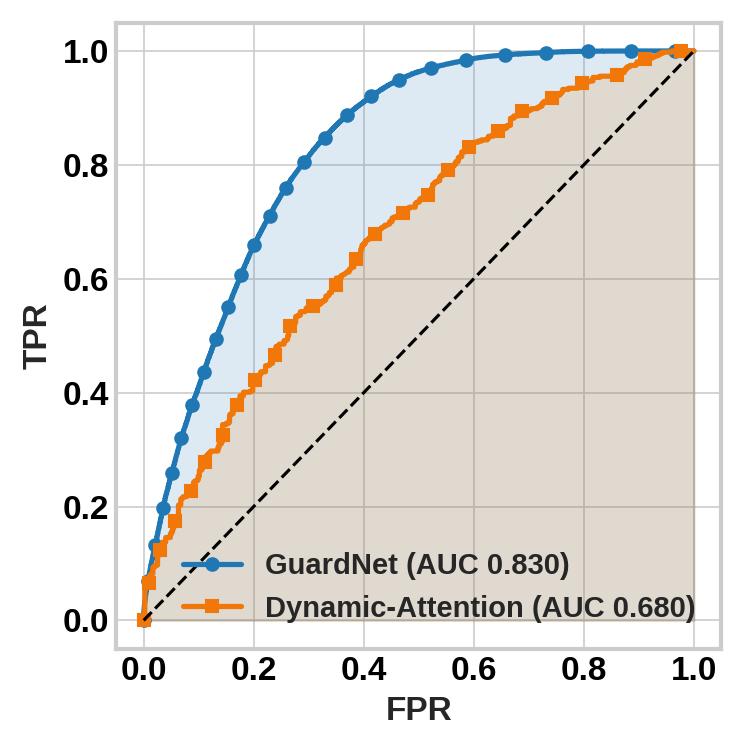}}\hfill
  \subfloat[PLeak – Financial]{%
    \includegraphics[width=0.32\textwidth]{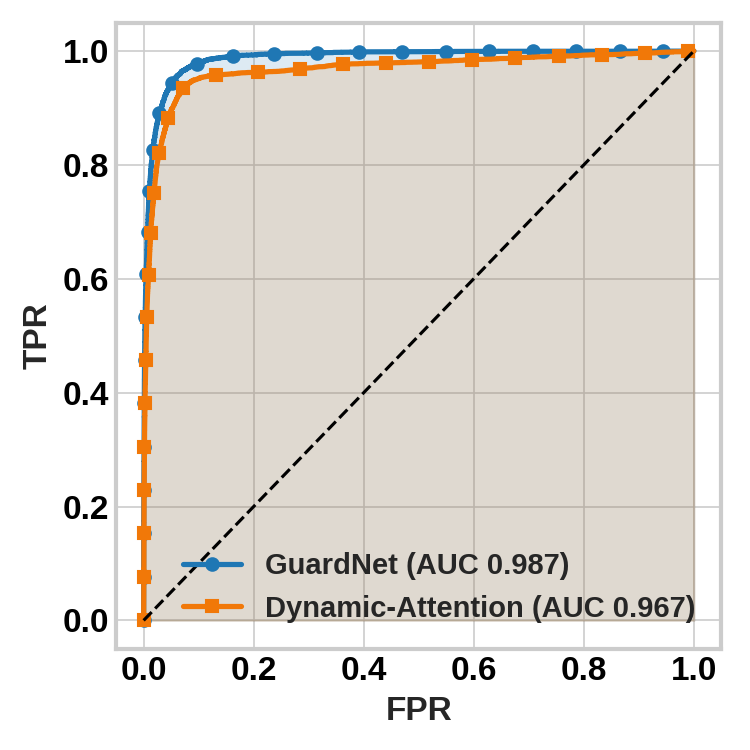}}\hfill
  \subfloat[PLeak – Tomatoes]{%
    \includegraphics[width=0.32\textwidth]{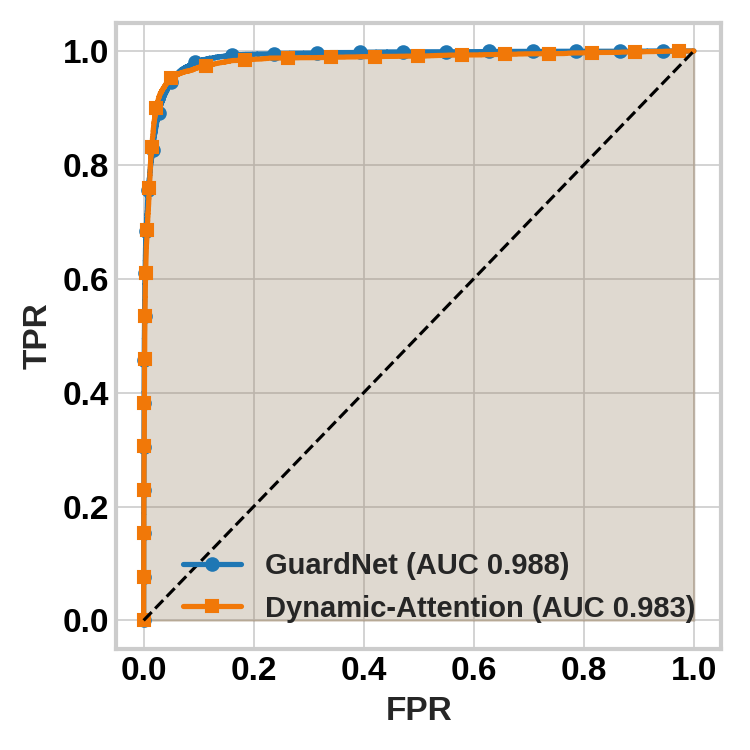}}
  \caption{ROC curves for GuardNet vs.\ baselines.
           Top row: prompt-level detection against TextDefense. Bottom row: token-level detection against Dynamic-Attention.}
  \label{fig:roc_all}
\end{figure*}

\paragraph{ROC Analysis (Figure~\ref{fig:roc_all})}
The receiver operating characteristic (ROC) curves visualize the discriminative ability of GuardNet and competing systems across decision thresholds.  
At the \emph{prompt-level} (top row), GuardNet maintains nearly perfect ROC profiles, with AUC scores of 100.0\% on LLM-Fuzzer, 99.8\% on \textsc{PLeak}–Financial, and 100.0\% on \textsc{PLeak}–Tomatoes.  
In contrast, TextDefense performs significantly worse, with AUCs of only 59.4\%, 67.1\%, and 80.7\% on the respective datasets. The discrepancy is especially stark on LLM-Fuzzer, where TextDefense’s ROC curve barely rises above the diagonal, indicating poor separation between harmful and benign prompts.

The performance gap is attributable to TextDefense’s reliance on entropy-based WISD scoring, which exhibits limited resolution. On domains like \textsc{PLeak}, with short and syntactically similar prompts, the resulting scores often lack granularity, compressing many examples into similar score bands.

At the \emph{token-level} (bottom row), GuardNet again outperforms Dynamic-Attention. It achieves AUCs of 83.0\% on LLM-Fuzzer, 98.7\% on \textsc{PLeak}–Financial, and 98.8\% on \textsc{PLeak}–Tomatoes.  
Dynamic-Attention trails behind at 68.0\%, 96.7\%, and 98.3\%, respectively. Although performance is closely matched on \textsc{PLeak}–Tomatoes, GuardNet demonstrates superior robustness under more adversarial prompt styles, as seen in LLM-Fuzzer.

Overall, the ROC analysis confirms that GuardNet offers the most favorable trade-off between true and false positives, especially in safety-critical contexts where precision and reliability are essential. For completeness, we also analyze precision–recall (PR) curves, which provide complementary insights under class imbalance; these results are reported in Appendix~\ref{sec:appendix}.

\section{Discussion}
\label{sec:discussion}

GuardNet provides a unified and robust framework for jailbreak detection, offering several key advantages over existing baselines, along with tradeoffs that depend on deployment needs.

\subsubsection{Advantages}
\begin{itemize}
    \item \textbf{Unified multi-level detection:}  
    In contrast to \emph{TextDefense}, which operates exclusively at the \emph{prompt-level}, and \emph{Dynamic Attention}, which functions solely at the \emph{token-level}, GuardNet delivers joint detection across both levels. This enables high-level filtering of risky prompts alongside fine-grained localization of adversarial spans.

    \item \textbf{Enhanced structural expressiveness:}  
    GuardNet constructs a heterogeneous graph over each input by combining syntactic edges with attention-derived dependencies. This allows it to model both local fluency and long-range semantic triggers. Prior baselines, relying on dispersion heuristics (\emph{TextDefense}) or statistical masking (\emph{Dynamic Attention}), are not equipped to capture such higher-order interactions.

    \item \textbf{Span-level interpretability:}  
    GuardNet outputs token-level binary masks identifying the specific regions responsible for adversarial behavior. While \emph{TextDefense} returns only a global anomaly score per prompt, and \emph{Dynamic Attention} lacks a mechanism for full-prompt detection, GuardNet offers interpretable outputs at both granularities, improving transparency and controllability.

    \item \textbf{Improved cross-domain generalization:}  
    GuardNet avoids reliance on static features or domain-tuned thresholds, instead leveraging patterns in self-attention and linguistic structure. This design yields significantly better transferability to unseen domains, where \emph{TextDefense} in particular shows degraded performance due to limited scoring resolution and prompt sensitivity.
\end{itemize}

\subsubsection{Limitations and tradeoffs}
\begin{itemize}
    \item \textbf{Computational cost:}  
    GuardNet incurs higher inference latency than lighter baselines, due to graph construction and two passes through a graph neural network.  

    \item \textbf{Hyperparameter sensitivity:}  
    As with competing methods namely, TextDefense (with sampling ratio, and decision threshold hyperparameters) and Dynamic‐Attention (with masking fraction, and $\beta$-scaling hyperparameters), GuardNet also requires hyperparameter tuning.  
    Key parameters include the number of top-$k$ attention neighbors, the local attention window size, and the focal-loss class weights.  
    This flexibility enables GuardNet to adapt effectively across different model scales and prompt lengths, but also introduces a broader hyperparameter search space.  
    Careful tuning and validation are therefore essential to fully realise GuardNet’s potential detection performance.

\end{itemize}

\section{Conclusion}
\label{sec:conclusion}

In this paper, we presented GuardNet, a hierarchical detection framework for proactively identifying and suppressing jailbreak prompts before model inference. By combining a prompt-level graph-based encoder with a token-level attention-guided detector, GuardNet captures both global structural patterns and fine-grained adversarial cues. 

Extensive experiments across multiple datasets and attack types demonstrate that GuardNet consistently outperforms state-of-the-art baselines, namely \emph{TextDefense} and \emph{Dynamic Attention}, in terms of accuracy, precision, recall, F1-score, and cross-domain robustness. Its two-level architecture ensures computational efficiency: the prompt-level filter quickly detects adversarial prompts, activating the token-level detector.

In future work we will explore optimizations such as lightweight graph construction, model quantization, and hardware-aware scheduling to reduce latency further. In addition, we aim to adapt GuardNet for deployment with on-device LLMs, enabling secure and privacy-preserving jailbreak detection without reliance on external servers. Finally, extending GuardNet to multilingual and multi-modal scenarios presents a promising direction toward more universally robust jailbreak detection systems.





%

\bibliographystyle{IEEEtran}
\bibliography{References}

\section*{Appendix}
\label{sec:appendix}
\begin{figure*}[t]
  \centering
  \subfloat[LLM-Fuzzer]{%
    \includegraphics[width=0.32\textwidth]{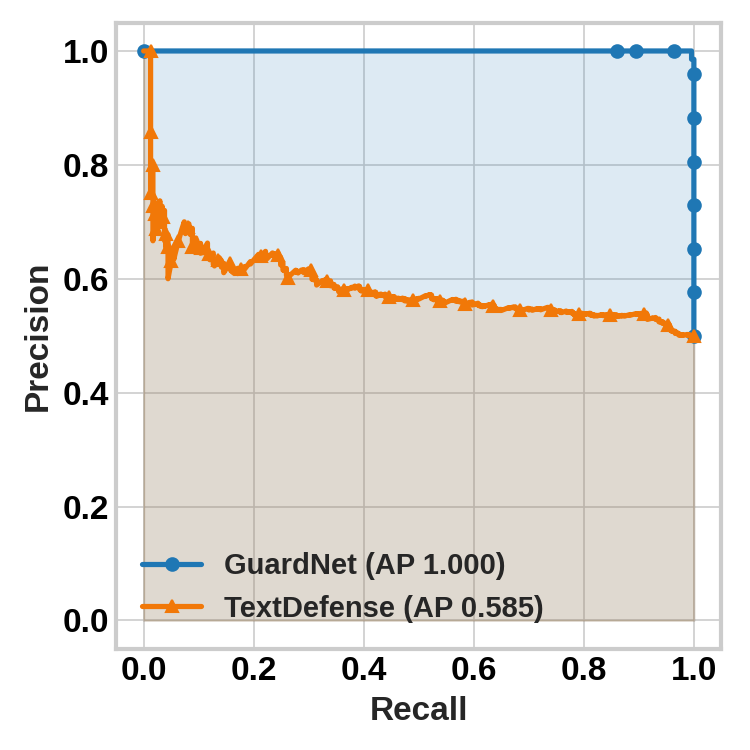}}\hfill
  \subfloat[PLeak – Financial]{%
    \includegraphics[width=0.32\textwidth]{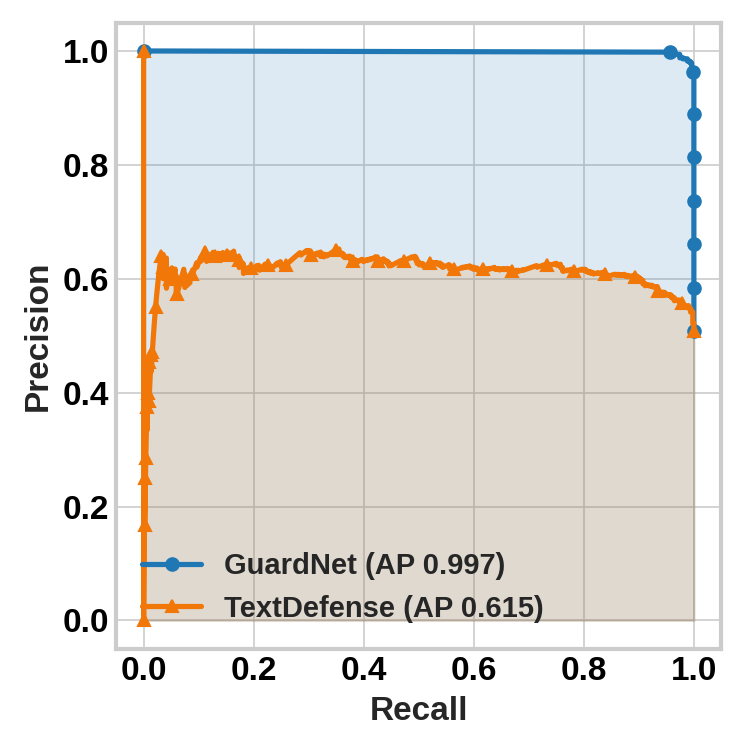}}\hfill
  \subfloat[PLeak – Tomatoes]{%
    \includegraphics[width=0.32\textwidth]{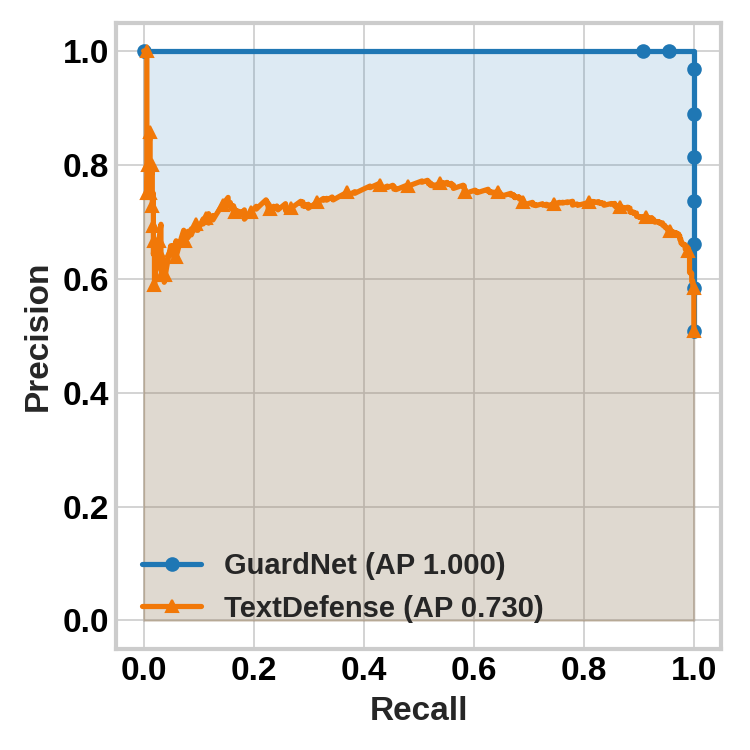}}\\[0.8em]
  \subfloat[LLM-Fuzzer]{%
    \includegraphics[width=0.32\textwidth]{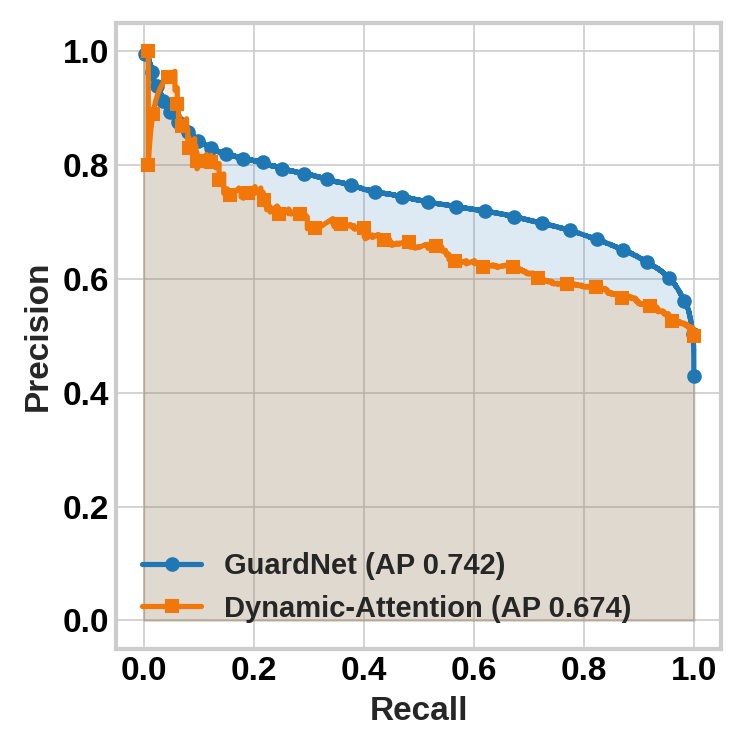}}\hfill
  \subfloat[PLeak – Financial]{%
    \includegraphics[width=0.32\textwidth]{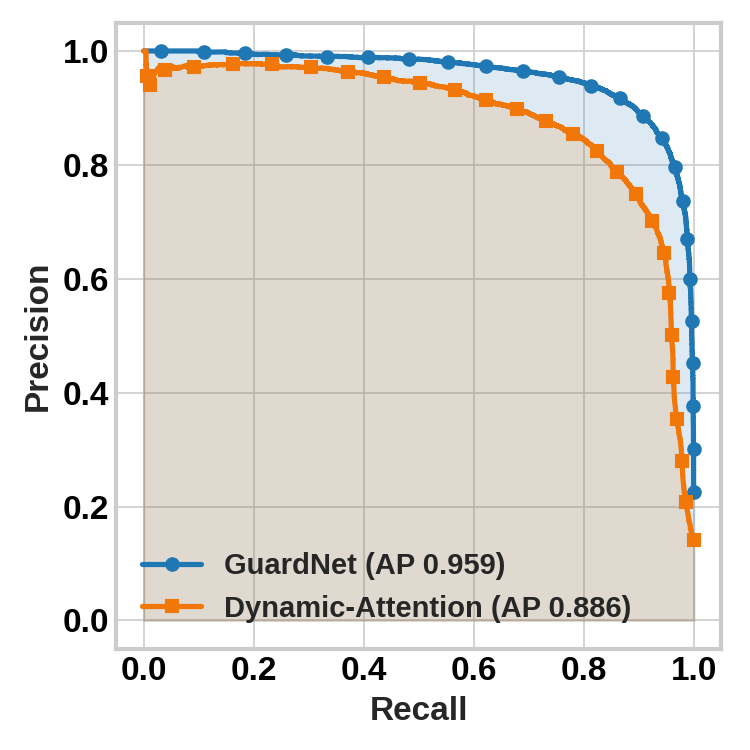}}\hfill
  \subfloat[PLeak – Tomatoes]{%
    \includegraphics[width=0.32\textwidth]{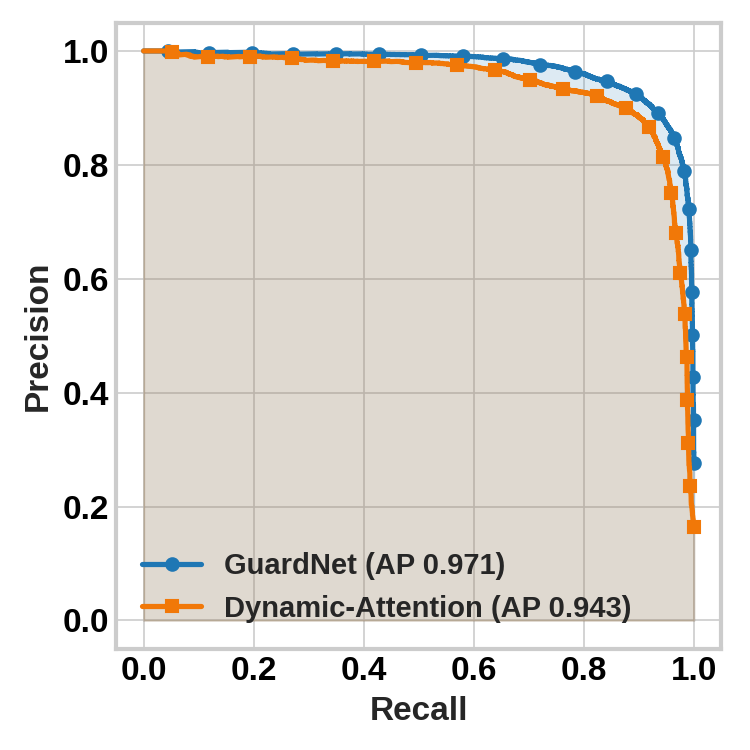}}
  \caption{Precision–Recall (PR) curves comparing GuardNet to baseline systems. Top row: prompt-level detection against TextDefense. Bottom row: token-level detection against Dynamic-Attention.}
  \label{fig:pr_all}
\end{figure*}

\paragraph{Precision–Recall Analysis (Figure~\ref{fig:pr_all})}
We further evaluate system performance using precision–recall (PR) curves, which provide complementary insight to ROC analysis and are particularly informative under class imbalance.  
At the \emph{prompt-level} (top row), GuardNet delivers near-perfect performance, with average precision (AP) scores of 100\% on both LLM-Fuzzer and \textsc{PLeak}–Tomatoes, and 99.7\% on \textsc{PLeak}–Financial.  
Its PR curves remain sharply concentrated in the upper-right corner, indicating that GuardNet can reliably retrieve adversarial prompts without sacrificing precision, even at high recall.

In contrast, TextDefense suffers from substantially reduced APs of 58.5\%, 61.5\%, and 73\% on the respective datasets. On LLM-Fuzzer in particular, TextDefense’s precision drops steeply after minimal recall, revealing a strong tendency toward over-flagging. While the PR curves appear relatively smooth, they remain low across the recall spectrum, reflecting the shallow discriminative capacity of entropy-based WISD scoring under diverse prompt conditions.

At the \emph{token-level} (bottom row), GuardNet again surpasses Dynamic-Attention. It achieves APs of 74.2\%, 95.9\%, and 97.1\% on LLM-Fuzzer, \textsc{PLeak}–Financial, and \textsc{PLeak}–Tomatoes, respectively.  
Dynamic-Attention trails slightly behind at 67.4\%, 88.6\%, and 94.3\%. While the performance gap narrows on \textsc{PLeak}–Tomatoes, GuardNet maintains a consistent advantage, especially at lower recall thresholds, where Dynamic-Attention's precision erodes more rapidly.

Overall, the PR analysis further validates GuardNet’s superior balance between precision and recall, particularly in settings where missing true positives is costly and over-alerting must be avoided.

\end{document}